\pdfoutput=1
\documentclass[11pt,a4paper,final]{article}
\usepackage[utf8]{inputenc}
\usepackage{algorithm}
\usepackage{algorithmic}
\usepackage{fullpage}

\usepackage{amsmath}
\usepackage{amsthm}
\usepackage{amssymb}
\usepackage{hyperref}
\usepackage{xcolor}
\usepackage{amsfonts}
\usepackage{mathtools}
\usepackage{apxproof}
\usepackage{multicol}
\usepackage[capitalize,noabbrev]{cleveref}
\usepackage{natbib}
\usepackage{authblk}
\newtheoremrep{theorem}{Theorem}

\newtheoremrep{lemma}[theorem]{Lemma}
\newtheorem{claim}[theorem]{Claim}
\newtheorem{definition}[theorem]{Definition}
\newtheorem{assumption}[theorem]{Assumption}

\newtheorem{proposition}[theorem]{Proposition}
\newtheorem{fact}[theorem]{Fact}
\newtheoremrep{corollary}[theorem]{Corollary}
\DeclareMathOperator{\Tr}{Tr}
\DeclareMathOperator*{\E}{\mathbb{E}}

 \usepackage[suppress]{color-edits}

\DeclareMathOperator{\argmax}{\text{arg}\max}

\newcommand{\eps}{\varepsilon}
\newcommand{\cA}{\mathcal{A}}

\newcommand{\cX}{\mathcal{X}}

\newcommand{\cC}{\mathcal{C}}

\newcommand{\cR}{\mathcal{R}}
\newcommand{\cY}{\mathcal{Y}}

\newcommand{\cZ}{\mathcal{Z}}
\newcommand{\cN}{\mathcal{N}}
\newcommand{\cS}{\mathcal{S}}
\newcommand{\cU}{\mathcal{U}}

\newcommand{\bbR}{\mathbb{R}}
\newcommand{\bbP}{\mathbb{P}}
\newcommand{\bbZ}{\mathbb{Z}}

\newcommand{\bbE}{\mathbb{E}}

\newcommand{\pr}[1]{\mathrm{Pr}\left[#1\right]}

\newcommand{\w}{\boldsymbol{w}} 
\newcommand{\y}{\boldsymbol{y}} 
\newcommand{\va}{\boldsymbol{a}} 
\newcommand{\x}{\boldsymbol{x}} 
\newcommand{\z}{\boldsymbol{z}} 
\newcommand{\etabf}{\boldsymbol{\eta}} 
\newcommand{\vb}{\boldsymbol{b}} 

\newcommand{\V}{\boldsymbol{V}} 
\newcommand{\I}{\boldsymbol{I}} 

\newcommand{\X}{\boldsymbol{X}} 
\newcommand{\M}{\boldsymbol{M}} 
\newcommand{\LambdaMAT}{\boldsymbol{\Lambda}} 

\newcommand{\phib}[1]{\phi\left( #1 \right)}

\newcommand{\Lambtil}{\widetilde{\LambdaMAT}}

\newcommand{\lambtil}{\widetilde{\lambda}}

\newcommand{\ytil}{\widetilde{\y}}

\newcommand{\BM}{\mbox{{\sf BinMech}}}

\newcommand{\pp}[1]{\left( #1 \right)} 
\newcommand{\nn}[1]{\left\| #1 \right\|} 
\newcommand{\brac}[1]{\left[ #1 \right]} 
\newcommand{\ii}[2]{^{#1}_{#2}}
\newcommand{\vdot}[1]{\left\langle #1 \right \rangle}

\newcommand{\defeq}{\vcentcolon=}

\newcommand{\Gau}{\textsc{Gau}}

\newcommand{\PCkth}[1]{\cC_{h}^{(#1)} (\tfrac{\rho}{2H})}
\newcommand{\PC}{\PCkth{\cdot }}

\newcommand{\alglinelabel}{%
  \addtocounter{ALC@line}{-1}
  \refstepcounter{ALC@line}
  \label
}

\newcommand{\ktil}{{\tilde{k}}}

\newcommand{\gaumechboundvar}{\lambtil_y}

\newcommand{\binmechboundvar}{\lambtil_\Lambda}

\newcommand{\switchcostvar}{N_{\text{max}}}

\newcommand{\switchcountvar}{N_{\text{count}}}

\newcommand{\switchingcost}{\frac{dH}{\log 2 } \log \left(1 + \frac{K}{\binmechboundvar d}  \right) }

\newcommand{\gaumechbound}{\sqrt{
\tfrac{
    \left(H \cdot\switchingcost\right)
    ^2 
     \log\left(\frac{3KH}{p}\right)
}{\rho}}}

\newcommand{\binmechbound}{\tfrac{\log(K) \left(
    6\sqrt{d+1} + 2 \log\left(\frac{3KH}{p}\right)
    \right)}{\sqrt{2\rho}}}
    
\newcommand{\betabound}{5 H^2 \sqrt{d\binmechboundvar\log(\chi)} + 6dH \sqrt{\log(\chi)}}

\newcommand{\chibound}{\frac{25^4 \cdot 162 \cdot K^4 \cdot d \cdot U_K \cdot H}{p}}

\addauthor{sw}{blue}
\addauthor{dn}{red}
\addauthor{gv}{green}

\title{Improved Regret for Differentially Private Exploration in Linear MDP}

\begin{document}
\author[1]{Dung Daniel Ngo\thanks{Indicates equal contribution.}}
\author[1]{Giuseppe Vietri$^*$}
\author[2]{Zhiwei Steven Wu}
\affil[2]{Carnegie Mellon University, \{zstevenwu\}@cmu.edu}
\affil[1]{University of Minnesota, \{ngo00054, vietr002\}@umn.edu}
\date{}
\maketitle

\begin{abstract}
We study privacy-preserving exploration in sequential decision-making for environments that rely on sensitive data such as medical records. In particular, we focus on solving the problem of reinforcement learning (RL) subject to the constraint of (joint) differential privacy in the linear MDP setting, where both  dynamics and rewards are given by linear functions. Prior work on this problem due to \cite{luyo2021differentially} achieves a regret rate that has a dependence of $O(K^{3/5})$ on the number of episodes $K$. We provide a private algorithm with an improved regret rate with an optimal dependence of $O(\sqrt{K})$ on the number of episodes. The key recipe for our stronger regret guarantee is the \emph{adaptivity} in the policy update schedule, in which an update only occurs when sufficient changes in the data are detected. As a result, our algorithm benefits from \emph{low switching cost} and only performs $O(\log(K))$ updates, which greatly reduces the amount of privacy noise. Finally, in the most prevalent privacy regimes where the privacy parameter $\epsilon$ is a constant, our algorithm incurs negligible privacy cost---in comparison with the existing non-private regret bounds, the additional regret due to privacy appears in lower-order terms. 

\end{abstract}

\section{Introduction}

 Many real world machine learning (ML) systems operate under the setting of \emph{interactive learning}, where learning algorithms interact with users and collect feedback from them over time. In domains such as personalized medicine where ML algorithms rely on sensitive data, the ability to protect users' data privacy has become increasingly critical. In order to provide rigorous and formal privacy guarantees, there has been a growing and longstanding literature on designing ML algorithms subject to the constraint of \emph{differential privacy} (DP) \citep{dwork2006calibrating}. While the vast majority of the DP ML literature has focused on the setting of supervised learning, there has been significantly less development on interactive learning, especially reinforcement learning (RL).

In this paper, we advance a recent line of work starting from \citet{vietri2020private} that provides RL algorithms with provable  privacy guarantees and performance bounds. Concretely, we consider a setting of episodic RL where an agent interacts with  $K$ users that arrive sequentially over $K$ episodes. In each episode $k$, the agent interacts with user $k$ over a fixed horizon of $H$ time steps. At each time step in the episode, the current user reveals their state to the agent, then the agent provides the user with a recommended action to take, which subsequently generates a reward received by the user. The goal of the agent is to maximize the cumulative reward over all users, or equivalently, minimize the regret with respect to the optimal policy. In this model, the sequence of states and rewards of each user is considered sensitive information. While each user may be willing to share such information to the agent in exchange for services or recommendations, there is still a risk that the agent inadvertently leaks the user's private information through the interactions with other users.

In order to prevent such privacy risks, existing work on private contextual bandits \cite{shariff2018differentially} and private RL \cite{vietri2020private} have adopted the notion of \emph{joint differential privacy} (JDP) \cite{kearns2014jdp}, a variant of DP that is suitable for sequential learning settings. Informally, JDP requires that for any user $k$, the output information to all other users except $k$ cannot reveal much about $k$'s private data. As a consequence, even if all other users collude (e.g., collectively probing the agent's policies) against $k$, the private information of $k$ is still protected.

Under the constraint of JDP, earlier work by \citet{vietri2020private, garcelon2020local} focuses on the tabular MDP setting, where the states and actions are discrete and the value functions can be stored in a table. More recent work has considered function approximation for DP RL. In particular, \citet{luyo2021differentially} considers the \emph{linear MDP setting}, in which the transition dynamics and reward function are assumed to be linear. All of these algorithms obtain the JDP guarantee under the so-called \emph{billboard model} \cite{hsu2016private}, in which the RL agent continuously and differentially privately releases a collection of statistics that are sufficient for computing each user's recommended actions when given their own private data. The key step to obtain low regret is to maintain these sufficient statistics at a low privacy costs. To that end, these algorithms in prior work leverage the \emph{binary mechanism} (and its variations)~\citep{panprivacy, chan2011private, shariff2018differentially}, which can continually release any form of summation or count statistics subject to DP with error only scaling logarithmically in the total number of episodes. 

While bringing function approximation to private RL greatly expands its scope and practicality, it also introduces new challenges that the existing privacy techniques based on the binary mechanism cannot handle. In particular, algorithms in the linear MDP setting typically maintain and update the value function $V$, which cannot be re-written as a form of summation statistics on the private data. This barrier has led to a sub-optimal regret in the prior work of \citep{luyo2021differentially}. Since their algorithm cannot leverage the binary mechanism for tracking the value functions privately, they resorts to applying the Gaussian mechanism repeatedly over an non-adaptive schedule, which incurs a privacy cost that scales with $K^{3/5}$ in their regret. Our work provides a private RL algorithm that circumvents this barrier and achieves a significantly lower privacy cost---in comparison to the non-private bounds, the privacy cost is  in lower order.

\subsection{Our contributions.}
We focus on  the setting of linear MDP, where there exists a feature map $\phi$ that maps each state-action pair $(x, a)$ to a $d$-dimensional vector $\phi(x, a)$. For each $(x, a)$, we assume that both the reward and transition function are linear in $\phi(x, a)$. Our contributions include the following:

\paragraph{Our results.}  We provide an $(\eps, \delta)$-JDP RL algorithm that achieves the state-of-the-art regret bound of
$$\widetilde{O} \pp{\sqrt{d^3 H^4 K} + \frac{\dnedit{H^3} d^{5/4} K^{1/2}}{\epsilon^{1/2}} },$$
The best known bound in prior work \citet{luyo2021differentially} is 
$\widetilde{O}\pp{\sqrt{d^3 H^4 K} + \frac{d^{8/5} H^{11/5} K^{3/5}}{\epsilon^{2/5}}}$. Note that both regret bounds include two terms, in which the first term corresponds to regret of a non-private algorithm and the second term corresponds to the "cost of privacy." In both results, the non-private term matches the state-of-the-art non-private regret bound from \citet{jin2020provably}. However,  our regret bound improves the cost of privacy from \citet{luyo2021differentially} in the parameters \dndelete{$H$,}$d$ and $K$. In the most prevalent regimes of differential privacy, the privacy parameter $\eps$ is chosen to be a small constant. In this case, the privacy cost term is dominated the non-private regret rate, and thus, our regret rate becomes $\widetilde{O} \pp{\sqrt{d^3 H^4 K}}$, matching the non-private rate from \citet{jin2020provably}. 

\paragraph{Our techniques.} The key technical ingredient that enables our improved regret bound is \emph{adaptivity}. In particular, our algorithm only updates its underlying policy when it detects a sufficient change in the collected data. Unlike prior work \cite{luyo2021differentially} that employs an non-adaptive update schedule and triggers polynomial in $K$ number of updates, our algorithm  draws the \emph{low switching cost} techniques from \citet{wang2021provably} and only triggers policy update roughly $O(\log(K))$ times. Low switching cost is particularly appealing for privacy since it largely reduce the amount of noise needed to achieve the same level of privacy parameters, which in turn leads to an improved regret. 

 The adaptive policy update schedule introduces challenges for the privacy analysis, since the policy update time is an unknown random variable that depends on the data and the randomness of the algorithm. In order to exploit the advantage of low switching cost, our privacy analysis needs to bound the privacy loss to be only proportional to the number updates, instead of total number of episodes $K$. To meet this challenge, our privacy analysis relies on a novel argument that models the interactions with the users as an adaptive adversary with bounded sensitivity. We believe that our low-switching-cost algorithm and analysis provide fruitful directions for answering other private RL questions in future, since they go beyond the existing paradigm that heavily rely on the binary mechanism.

\subsection{Related Work}


Our work is most related to the line of work on RL subject to JDP that includes the earlier work of \citet{vietri2020private,garcelon2020local} in the tabular settings and more recent work of \cite{luyo2021differentially} in the linear MDP settings. Both \cite{Zhou22} and \cite{luyo2021differentially} also consider a different form of linear function approximation called the linear mixture MDPs, which we do not study in this work. The linear mixture MDPs setting is arguably easier to learn under JDP since all the relevant sufficient statistics for learning can be written as summation and the binary mechanism is applicable. In contrast, obtaining an $O(\sqrt{K})$ regret rate in our linear MDP setting requires new techniques.

More broadly speaking, our work contributes to the growing line of work on private interactive learning, which includes the study of online learning \citep{TS13, agarwal2017price}, multi-armed bandits \citep{mishra2015mab, Tossou2017AchievingPI}, linear contextual bandits \citep{shariff2018differentially}.

Technically, our algorithm is closely related to the non-private algorithms in the linear MDP settings that achives provable low regret \cite{jin2020provably}. Low switching cost has also been a desideratum for RL algorithms, even absent privacy concerns.  For example,  \cite{bai2019provably} shows how to bound local switching cost in a tabular MDP setting: $O(H^3SA\log(K))$, and \cite{gao2021provably, wang2021provably} give a provably efficient algorithm for linear MDP with low switching cost.

\section{Preliminaries}

\paragraph{Notation}
We use bold capital letters to denote matrices, bold lower case for vectors. 
Let $\X_{1:t}$ be the matrix whose rows are $\x_1,\ldots,\x_t$,
then we define the Gram matrix by $\LambdaMAT_t = \X_{1:t}^\top \X_{1:t} = \sum_{i=1}^t \x_i\x_i^\top$. A symmetric matrix $\X$ is positive-semidefinite if $\x^\top \X \x \geq 0$ for any vector $\x$. Any such $\X$ defines a norm on vectors, so we define $\nn{\x}^2_{\X} = \x^\top \X \x$.
For any positive integer $N$, we use $[N]$ to denote the set $\{1, \ldots , N\}$.
%
We use $\z\sim \cN(0, \sigma^2)^d$  to denote a vector-valued random variable of dimension $d$, where each coordinate is sampled i.i.d from a Gaussian distribution with variance $\sigma^2$. For a vector $\x \in \mathbb{R}^d$, we use $\nn{\x}_2 = \sqrt{\sum_{i=1}^d \x_i^2}$ to denote the $\ell_2$ norm of $\x$. We can also define the spectral norm of a matrix $\M$ as $\nn{\M} = \max_{\x \neq 0} \frac{\nn{\M \x}_2}{\nn{\x}_2}$, which is the operator norm associated with the vector $\ell_2$ norm.

\subsection{Markov Decision Process}

We begin with the general setup for the episodic \emph{Markov Decision Process} (MDP) before we describe the linear MDP setting we focus on. An MDP is denoted by $\text{MDP}(\cS, \cA, H, \bbP, r)$, where $\cS$ is the set of states, $\cA$ is the set of actions, $H\in \bbZ_{+}$ is the length of each episode, $\bbP = \{\bbP_h\}_{h=1}^H$ are the state transition probability measures, and finally $r=\{r_h\}_{h=1}^H$ is the set of reward functions. We assume that $\cS$ is a measurable space with possibly infinite number of elements and $\cA$ is a finite set with cardinality $A$. For each $h\in H$, $\bbP_h(\cdot| x, a)$ denoted the transition kernel over next states if  action $a$ is taken for state $x$ at step $h\in [H]$. Similarly, at each round $h$, we denote the deterministic reward function as $r_h(x_h, a_h)$ in $[0, 1]$.\footnote{For notational simplicity, we study deterministic rewards, and our results can be generalized to random reward functions.} 
In the first time step of each episode, the agent observes the initial state $x_1$, which can be adversarially selected. At each time-step $h \leq H$, the agent observes $x_h\in \cS$, then picks an action $a_h\in \cA$ and receives the reward $r_h(x_h, a_h)$. As a result, the MDP moves to a new state $x_{h+1}$ drawn from probability measures $\mathbb{P}_h(x_{h+1}|x_h, a_h)$. When state $x_{H+1}$ is reached, the episode ends and the agent receives no further reward.

A deterministic policy $\pi$ is a function $\pi: \mathcal{S} \times [H] \rightarrow \mathcal{A}$, where $\pi(x_h, h)$ is the action that the agent takes at state $x_h$ and round $h$ in the episode. For any $h \in [H]$, we define the value function $V_h^{\pi}: \mathcal{S} \rightarrow \mathbb{R}$ as the expected value of cumulative rewards by following policy $\pi$ from round $h$ at an arbitrary state: 
\begin{equation*}
    V_h^{\pi}(x) \coloneqq \mathbb{E} \left[ \sum_{h'=h}^H r_{h'}(x_{h'}, \pi(x_{h'}, h')) \Big| x_h = x \right] \tag{$\forall x \in \mathcal{S}, h \in [H]$}
\end{equation*}

Moreover, we also define the action-value function $Q_h^{\pi}: \mathcal{S} \times \mathcal{A} \rightarrow \mathbb{R}$ to be the expected value of cumulative rewards when the agent starts from an arbitrary state-action pair at the $h$-th step and follow policy $\pi$: 
\begin{align*}
    Q_h^{\pi}(x,a) 
    \coloneqq r_h(x,a)  + 
    \mathbb{E} \left[ \sum_{h' = h+1}^H r_{h'}(x_{h'}, \pi(x_{h'}, h')) \Big|x_h = x, a_h = a \right] \tag{$\forall (x,a) \in \mathcal{S} \times \mathcal{A}, \forall h \in [H]$}    
\end{align*}

 To simplify notation, we denote the expected value function as $\brac{\bbP_h V_{h+1}}(x,a) = \bbE_{x'\sim \bbP_h(.|x,a)}{V_{h+1}(x')}$. The \emph{Bellman equation} then becomes 
\begin{align*}
    Q_h^\pi(x,a) &\defeq  \brac{r_h + \bbP_h V^\pi_{h+1}}(x, a)\\
    \quad V_{h}^\pi(x) &\defeq Q_h^\pi(x,\pi(x, h))\\
    \quad V_{H+1}^\pi(x) &\defeq 0 \text{ for all } x\in \cS
\end{align*}
which holds for all $(x,a) \in \cS\times \cA$. Similarly, the optimal Bellman equation is 
\begin{align*}
    Q_h^\star(x,a) &\defeq  \brac{r_h + \bbP_h V^\star_{h+1}}(x, a)\\
    \quad V_{h}^\star(x) &\defeq \max_{a\in \cA}Q_h^\star(x,a)\\ 
    \quad V_{H+1}^\star(x) &\defeq 0 \text{ for all } x\in \cS
\end{align*}

The optimal policy $\pi^\star$ is a greedy policy with respect to the optimal action-value policies $\{Q_h^\star\}_{h\in [H]}$ and achieves value $\{V_h^\star\}_{h\in [H]}$.


 The agent interacts with the environment over a sequence of $K$ episodes. At the beginning of each episode $k\geq 1$, the environment chooses an initial state $x_1^k$ and the agent chooses a policy $\pi_k$. Then, the agent's expected regret for the $k$-th episode is the difference in the values $V_1^\star(x_1^k) - V_1^{\pi_k}(x_1^k)$. Therefore, after playing for $K$ episodes the total expected regret is
\begin{align*}
    \text{Regret}(K) = \sum_{k=1}^K V_1^\star(x_1^k) - V_1^{\pi_k} (x_1^k)
\end{align*}

\subsection{Linear MDP}

We focus on linear \textit{Markov Decision Process} setting, where reward and transition functions are assumed to be linear \citep{BradtkeB96, MeloR07}.

\begin{assumption}[Linear MDP]$\text{MDP}(\cS, \cA, H, \bbP, r)$ is a linear MDP with a feature map $\phi: \cS\times \cA\rightarrow \bbR^d$, if for any $h\in [H]$ there exist $d$ \textit{unknown } measures $\mu_h = \pp{\mu_h^1, \ldots, \mu_h^d}$ over $\cS$ and an \textit{unknown} vector $\theta_h$, such that for any $(x,a)\in \cA\times \cS$ we have 
\begin{align*}
\bbP_h(\cdot | x,a) = \vdot{\phi(x,a), \mu_h(\cdot )}, \quad
    r_h(x,a) = \vdot{\phi(x,a), \theta_h}
\end{align*}
We assume that $\| \phi(x,a)\| \leq 1 $ for all $(x,a)\in \cS\times \cA$, and $\max\{\|\mu_h(\cS)\|, \|\theta_h\|\}\leq d$ for all $h\in [H]$.
\end{assumption}


An important property of linear MDP is that the action-value functions are linear in the feature map $\phi$. Hence, it suffices to focus on linear action-value functions.

\begin{proposition}[\citet{jin2020provably}]
\label{prop:linear-func}
For a linear MDP, for any policy $\pi$, there exist weights $\{w^{\pi}_h\}_{h \in [H]}$ such that for any tuple $(x, a, h) \in \mathcal{S} \times \mathcal{A} \times [H]$, we have $Q^{\pi}_h (x, a) = \langle \phi(x,a), w^{\pi}_h \rangle$
\end{proposition}

\subsection{(Joint) Differential Privacy in Episodic RL}

In this work, we provide privacy-preserving RL algorithm that incorporates the rigorous notion of \emph{differential privacy} (DP). We first revisit the standard DP definition. Formally, we use $U \in \cU^K$ to denote a sequence of $K$ users participating in the RL protocol. Technically speaking, a user can be identified with a tree of depth $H$ encoding the state and reward responses they would give to all $A^H$ possible sequences of actions the agent can choose. We also say two users sequences $U$ and $U'$ are neighbors if they differ in just one user $i \leq k$.
\begin{definition}[DP \citep{dwork2006calibrating}] \label{def:DP}
A randomized mechanism $M$ satisfies $(\epsilon,\delta)$-differential privacy ($\pp{\epsilon, \delta}$-DP) if for all neighboring datasets $U,U'$ that differ by one record and for all event $E$ in the output range,
\begin{align*}
    \bbP\brac{ M(U) \in E} \leq e^\epsilon \bbP\brac{ M(U') \in E} + \delta
\end{align*}
When $\delta=0$ we say \textit{pure}-DP, and when $\delta>0$, then we say \textit{approximate}-DP.
\end{definition}

In our episodic MDP setting, there is a set of $K$ users arriving sequentially, and each user's sensitive data corresponds to the sequence of states and rewards in a single episode. We focus on the \emph{central model} where users are willing to share their sensitive information with a trusted agent in exchange for a service or recommendation but they don't want their information to be leaked to third parties. Our goal is to avoid any inference about the user's information while interacting with the RL agent. The standard definition of DP is too stringent for this setting, since it would require the entire output, which includes the action recommendations for user in episode $k$, and cannot reveal user $k$'s private data, which prevents any utility in the recommendations. Following the prior work of \citet{vietri2020private}, we consider the notion of \emph{joint differential privacy} (JDP) \cite{KPRU}, which informally requires that if any single user changes their data, the information observed by all the other $(K - 1)$
users cannot not change substantially.

To introduce JDP formally, we denote $U=\pp{u_1, \ldots, u_K}$ the user sequence that interacts with the agent over $K$ episodes. Technically speaking, a user can be identified with a tree of depth $H$ encoding the state and reward responses they would give to all the possible sequences of actions the agent can choose. In their interaction, the agent only gets to observe the information along a single root-to-leaf path in each user’s tree. Let $\cU$ denote the space of all such trees. Let $M$ be any RL algorithm that takes as input a sequence of users $U$ and outputs action recommendations over $K$ episodes. For any $k\in[K]$ we denote $M_{-k}(U)$ as all action recommendations generated by $M$ except the action during the $k$-th episode. Two user sequences $U$ and $U'$ are $k$-neighbors if they only differ in their $k$-th user.

\begin{definition}\label{def:jointzCDP} (Joint-Differential Privacy (JDP))
An RL algorithm $M: \mathcal{U}^K \rightarrow \mathcal{A}^{H \times [K-1]}$ is $(\epsilon, \delta)$-jointly differentially private ($(\epsilon, \delta)$-JDP)  if for all $k \in [K]$, all $k$-neighboring user sequences $U, U'$, and all events $E \subseteq \mathcal{A}^{H \times [K-1]}$,
\begin{align*}
        \bbP\brac{ M_{-k}(U)\in E} \leq e^\epsilon \bbP\brac{ M_{-k}(U') \in E} + \delta
\end{align*}
\end{definition}
While we state our main results in terms of JDP, we will also use zero-Concentrated DP (zCDP) as a tool in our analysis, since it enables cleaner analysis for privacy composition and the Gaussian mechanism.

\begin{definition}[zCDP \citep{bun2016concentrated}] \label{def:CDP}
A randomized mechanism $M$ satisfies $\rho$-Zero-Concentrated Differential Privacy ($\rho$-zCDP), if for all neighboring dataset $U,U'$ and all $\alpha\in (1,\infty)$,
\begin{align*}
    D_\alpha (M(U) \parallel M(U')) \leq \rho \alpha
\end{align*}
where $D_\alpha$ is the Renyi-divergence 
\end{definition}

Any algorithm that satisfies $\rho$-zCDP also satisfies approximate-DP. The following proposition from \citet{bun2016concentrated} shows how to do the mapping between zCDP and approximate-DP.


\begin{lemma}\label{lem:cdptodp}
If $M$ satisfies $\rho$-zCDP then $M$ satisfies $\pp{\rho + 2\sqrt{\rho \log(1/\delta)}, \delta}$-DP.
\end{lemma}

Another basic but important property of zCDP is easy composition of zCDP mechanisms:
\begin{lemma}[zCDP Composition]
\label{lem:zcdpcomposition}
Let $M:\cU^K\rightarrow \cY$ and $M':\cU^K\rightarrow \cZ$ be randomized mechanisms. Suppose that $M$ satisfies $\rho$-zCDP and $M'$ satisfies $\rho'$-zCDP. Define $M'':\cU^K\rightarrow \cY \times \cZ$ by $M''(U) = \pp{M(U), M'(U)}$. Then $M''$ satisfies $(\rho+ \rho')$-zCDP.
\end{lemma}

To apply DP techniques to some mechanism we must know the sensitivity of the function we want to release. Here we give the definition and the notation we use:
\begin{definition}[$\ell_2$-Sensitivity]
\label{def:sensitivity}
Let $U \sim U'$ denote neighboring datasets. Then the $\ell_2$-sensitivity of a function $f: \cU \rightarrow \mathbb{R}^d$ is 
\begin{align*}
 \Delta\pp{f}  \defeq  \max_{U \sim U'}\| f(U) - f(U')\|_2 
\end{align*}
if $f:\cU^K \rightarrow \mathbb{R}^{d\times d}$ is a matrix valued function, then $\Delta\pp{f}  \defeq  \max_{U \sim U'}\| f(U) - f(U')\|_{op} $.
\end{definition}

In our analysis, we use the Gaussian mechanism:
\begin{definition}[Gaussian mechanism]
The  Gaussian mechanism 
$\Gau(U, f, \rho)$ takes as input a dataset $U$, 
a function $f:\cX\rightarrow \mathbb{R}^{d}$ and a privacy parameter 
$\rho>0$ and outputs
$    f(U) + \etabf$
where  each coordinate $i\in[d]$ is sampled as $ \etabf_i \sim \cN(0, \sigma^2)$ and $\sigma^2=\frac{\Delta(f)^2}{2\rho}$.
\end{definition}

We use the following facts about the Gaussian mechanism:
\begin{lemma}
For any dataset $U\in \cU^K$ and low sensitive function $f:\cU^K\rightarrow \mathbb{R}^d$ and privacy parameter $\rho$. 
The Gaussian mechanism satisfies $\rho$-zCDP and, for $\gamma\in(0,1)$, the error is given by:
\begin{align*}
    \mathbb{P}
    \pp{
    \|f(U) - \Gau(U, f, \rho)\|)_2 > \sqrt{\tfrac{d \Delta^2 \log(d/\gamma)}{\rho}}} < \gamma
\end{align*}
\end{lemma}




\section{JDP RL with Low Switching Cost}\label{sec:algorithm}

We now introduce our RL algorithm for the linear MDP setting. We will first revisit the non-private algorithm of Least-Squares Value Iterations (LSVI) and then introduce our techniques to make such an algorithm private with a desirable privacy-accuracy trade-off.

\paragraph{LSVI}

In the linear MDP setting, we leverage the fact that the action value function $Q_h^\star$ is linear in the feature vector (Proposition \ref{prop:linear-func}) and thus they can be estimated using the idea of Least-Squares Value Iterations \citep{bradtke1996linear, osband2016generalization}. In particular, the true action-value function $Q_h^\star$ is parameterized by a vector $\w_h^\star$. If we updated the policy on episode $k$, then \cref{alg:privrl} recursively estimates $\w_h^\star$ for all $h$ using LSVI. Let $Q_h^k$ and $\w_h^k$ denote the running estimates of the action-value functions and their corresponding parameters over episodes.
In each episode $k$, the algorithm computes the following sufficient statistics relevant to the estimation of $\w_h^\star$: For some $\lambda>0$, starting with $h = H$ down to $h=1$, compute
a least-squares estimate $\w_h^k$ via
\begin{align}
\label{eq:lambdadef}
\LambdaMAT_h^k &=  \lambda\I  + {\sum_{i=1}^{k-1} \phib{x_h^i,a_h^i}\phib{x_h^i,a_h^i}^\top
    } \\
\label{eq:ydef}
    \y_h^k &= {\sum_{i=1}^{k-1} \phib{x_h^i,a_h^i} \pp{r_h^i + V_{h+1}^{k}(x_{h+1}^i)}}\\
    \w_h^k &= \left(\LambdaMAT_h^k\right)^{-1}  \y_h^k,
\end{align}
which subsequently define the  the corresponding optimistic estimates for action-value and state-value functions:
\begin{align}
Q_h^k(x, a) &= \langle \phi(x, a), \w_h^k\rangle +  \beta\|\phib{\cdot, \cdot}\|_{\pp{\LambdaMAT_h^k}^{-1}}\\
V_h^k(x) &= \max_a Q_h^k(x, a)
\end{align}
Given these optimistic estimates, the non-private algorithm follows a greedy policy with respect to the estimates of action-value functions. 

%


\paragraph{Private LSVI-UCB} 
To provide a JDP variant of the LSVI algorithm, our strategy is to identify a collection of statistics that are compatible with differentially private releases and sufficient for constructing estimates for the action value functions.



\paragraph{Binary Mechanism} We first introduce the mechanism to privatize the sequence of Gram matrices $\LambdaMAT_h^k$, defined in \eqref{eq:lambdadef}. First, notice that the Gram matrix statistic is in form of a sum. This allows us to use a variation of the binary  mechanism due to \citet{shariff2018differentially}, which is a tree-based aggregation mechanism that sequentially releases sums of matrices privately.  Our goal is to use the binary mechanism to produces a sequence of privatized gram matrices, which we denote by $\{\Lambtil_h^k\}_{k,h\in[K]\times[H]}$.

%
%
\paragraph{Gaussian mechanism with low switching cost}
Our second set of statistics is $\y_h^k$ (defined in \cref{eq:ydef}). The challenge with privatizing $\y_h^k$ is that it does not take the form of summation, as it depends on the value function estimate $V_{h+1}^k$ for episode $k$.  
As a result, we cannot apply the binary mechanism. Therefore, we use the Gaussian mechanism on each update episode to privatize $\ytil_h^k$. However, adding Gaussian noise to every episode would lead to too much noise. For this reason, we minimize the number of episodes we update the policy. 

In \cref{alg:privrl}, an update episode occurs every time the determinant of the noisy gram matrix $\Lambtil_h^k$ doubles for any $h \in [H]$. Since \cref{alg:privrl} does not update the action-value function every round, we use $\tilde{k}$ to denote the last episode the policy $\pi_h^k$ was updated. 

As shown in \cite{wang2021provably}, this adaptive update schedule allows us to bound the number of policy updates by $O(dH\log(K))$ while increasing the regret only by a factor of $2$. Therefore, instead of applying composition over $K$ Gaussian mechanisms, we only have $O(dH\log(K))$. Having $O(\log(K))$ updates is critical for obtaining our improved regret rate. The algorithm from \cite{luyo2021differentially} uses a non-adaptive batching technique that leads to a number of updates polynomial in $K$. In the next section, we show how to analyze the privacy guarantee of this adaptive update algorithm.

\newcommand{\upt}{\textrm{UPDATE}}

\newcommand{\phixa}{\phi_{x, a}}
\newcommand{\phiih}{{\phi_{h}^{i}}}

\begin{algorithm}[h]
\begin{algorithmic}[1]
\INPUT Privacy parameter $\rho$, policy update rate $C$, fail probability $p$, confidence width $\beta$. 
\STATE \textbf{Notation:} Let $\phi(x_h^i, a_h^i) = \phiih$

\STATE \textbf{Set:} $\binmechboundvar \defeq \binmechbound$, and $\switchcostvar \defeq \switchingcost$ 

\STATE \textbf{Initialize:} For all $h\in[H]$ : 
$\Lambtil_h^1 \leftarrow 2 \binmechboundvar \I$, and a counter 
$\PC$ as in \cref{eq:bincounter}.
\STATE \textbf{Initialize:} $\tilde{k} \leftarrow 1 $, and  $\switchcountvar \leftarrow 1$. 


\FOR{$k=1, \dots, K$}

        \IF{$\exists h \in [H],{ \det\pp{\Lambtil_h^{k}}}/{\det\pp{\Lambtil_h^{\tilde{k}}}} \geq C $ \algorithmicand $\switchcountvar < \switchcostvar$} \alglinelabel{lin:update}
        \FOR{$h = H, \cdots, 1$} 

        \STATE $\y_h^k \leftarrow 
         \sum_{i=1}^{k-1} \phiih  (r_h^i + V_{h+1}^k(x_{h+1}^i) )$
         
        \STATE $\ytil_h^k \sim  \cN\pp{\y_h^k, \frac{6 H^2 \switchcostvar}{\rho}\I}^d$
        
        
        
        

        \STATE $\w_h^k \leftarrow \left( \Lambtil_h^{k} \right)^{-1} \ytil_h^k$
        
        \STATE $\text{B}_h^k(\cdot, \cdot) = \beta \left[ \phi(\cdot, \cdot)^\top ( \Lambtil_h^k )^{-1} \phi(\cdot, \cdot) \right]^{\tfrac{1}{2}}$ \alglinelabel{lin:bonus}       
       
        \STATE  $Q_h^k(\cdot, \cdot) 
        = \min\{\phi(\cdot,\cdot)^\top \w_h^k + \text{B}_h^k(\cdot, \cdot), H\} $

        

        
        \STATE Update $\tilde{k} \leftarrow k$ \algorithmicand $\switchcountvar \leftarrow \switchcountvar + 1$
        \ENDFOR
        \ENDIF

    \FOR{step $h=1\ldots, H$}
        \STATE Observe $x_h^k$
        \STATE Take action $a\ii{k}{h} \leftarrow \argmax_{a\in \cA} Q\ii{\tilde{k}}{h}\left(x\ii{k}{h}, a\right)$.
        \STATE Observe reward $r_{h}^k \leftarrow r_h(x_h^k, a_h^k)$ .
         \STATE Update  $\PC$ with  $\phi_h^k (\phi_h^k)^\top$ and set $ \Lambtil_h^{k+1} \leftarrow \PCkth{k} + 2\binmechboundvar\I$ \alglinelabel{lin:shiftgram}
    \ENDFOR
    \ENDFOR
\caption{Private LSVI-UCB}\label{alg:privrl}
\end{algorithmic}
\end{algorithm}

%






\section{Analysis}
We begin by analyzing the switching cost of \cref{alg:privrl}, then the privacy and regret guarantees.

\subsection{Switching Cost}\label{sec:switchingcost}
We begin by analysing the switching cost of \cref{alg:privrl}, which will be used in \cref{sec:privacy} for the privacy analysis .

\begin{theorem} \label{thm:switchingcost}

[Similar to Lemma C.3 in \cite{wang2021provably}]
Let $\binmechboundvar$ be defined as in \cref{alg:privrl}.
Condition on the event that 
$\nn{\Lambtil_h^k - \LambdaMAT_h^k}_{op} \leq \binmechboundvar$ for all $h, k \in [H] \times [K]$. 
For $C=2$ and $\binmechboundvar > 0$, the global switching cost of \Cref{alg:privrl} is bounded by: 
    $\switchcountvar \leq \frac{dH}{\log 2 } \log \left(1 + \frac{K}{\binmechboundvar d}  \right) $.
\end{theorem}
\begin{proof}[Proof Sketch]
See the full proof derivation in \Cref{sec:switching-cost-appendix}. We use a determinant-based analysis, similar to that of the non-private algorithm in \cite{wang2021provably}. The main difference in our proof is that the determinant of $\Lambtil_h^{\tilde{k}}$ is now at most $(\binmechboundvar + (k-1)/d)^d$ to account for the perturbation from binary mechanism. 
Also, we notice that every time the update criteria is met, the determinant of $\Lambtil_h^{k+1}$ increases by at most twice. Hence, in total we can bound the number of updates by $O \left(\frac{dH}{\log 2} \log(1 + \frac{K}{\binmechboundvar d}) \right)$.\end{proof}

The low switching cost guarantee above is crucial for our algorithm to achieve our improved regret guarantee since it allows us to reduce the level of noise to preserve privacy.

\subsection{Privacy Analysis}\label{sec:privacy}



This section provides the privacy analysis of \cref{alg:privrl}. But, first, we state our main privacy guarantee: 
\begin{theorem}[Privacy]\label{thm:jzcdp}
\Cref{alg:privrl} satisfies $(\epsilon, \delta)$-JDP.
\end{theorem}

For the privacy analysis of \cref{alg:privrl}, we use  zCDP \citep{bun2016concentrated} because it simplifies the composition of Gaussian mechanisms. At the end, we translate our results in terms of approximate DP. In this section, we show that releasing statistics  $\{\Lambtil_h^k\}_{k,h\in[K]\times[H]}$  satisfies $(\rho/2)$-zCDP and then show that releasing $\{\ytil_h^{\tilde{k}}\}_{k,h\in[K]\times[H]}$ also satisfies $(\rho/2)$-zCDP.
Thus, by composition of zCDP mechanisms (\cref{lem:zcdpcomposition}), we have that releasing the sufficient statistics satisfies $\rho$-zCDP and, by \cref{lem:cdptodp},  also satisfies $(\epsilon, \delta)$-DP  for $\epsilon = \rho + \sqrt{\rho\log(1/\delta)}$. 

It follows that \cref{alg:privrl} is $\pp{\epsilon, \delta}$-JDP by the Billboard lemma due to \citet{hsu2016private}, since the actions sent to each user depends on a function constructed with DP and their private data only. We state the billboard lemma here:

\begin{lemma}[Billboard lemma \citep{hsu2016private}]
Suppose that randomized mechanism $M: \cU \rightarrow \cR$ is $(\epsilon,\delta)$-differentially private. 
Let $U\in\cU$ be a dataset containing $n$ users.
Then, consider any set of functions $f_i:\cU_i\times \cR \rightarrow \cR_i$, for $i\in [n]$, where $\cU_i$ is the portion of the database containing the $i$'s user data. Then the composition $\{f_i\left(\mathrm{proj}_i(U), M(U))\right\}_{i\in [n]}$ is $(\epsilon, \delta)$-JDP, where $\mathrm{proj}:\cU\rightarrow \cU_i$ is the projection to $i$'s data. 

\end{lemma}






{We start with bounding the sensitivity of the two types of sufficient statistics.}

\begin{lemma}[$\ell_2$-sensitivity] 
\label{lem:sensitivity}
For any  $(k, h) \in[K]\times [H]$.
%
We have $\Delta\pp{\LambdaMAT_h^{k}} \leq 2$ and $\Delta\pp{\y_h^k} \leq 2H + 2$.
\end{lemma}

\begin{proof}[Proof Sketch]
In order to bound the sensitivity of $\y_h^k$, we need to consider any two neighboring user sequences $U$ and $U'$ with outputs $\y_h^k$ and $(\y')_h^k$, respectively. Let $i \leq k$ be some episode with $x_h^i \neq {x'}_h^i$ and $a_h^i \neq {a'}_h^i$, where $(x_h^i,a_h^i) \in U$ and $({x'}_h^i,{a'}_h^i) \in U'$.  By definition of neighboring user sequences and triangle inequality, the difference between $\y_h^k$ and $(\y')_h^k$ can be written as the sum of $\nn{\phi(x_h^i, a_h^i) \left(r_h(x_h^i, a_h^i) + V_{h+1}^k(x_{h+1}^i) \right)}_2$ and $\nn{\phi({x'}_h^i, {a'}_h^i) \left(r_h({x'}_h^i, {a'}_h^i) + V_{h+1}^i({x'}_{h+1}^i) \right)}_2$. Also, by assumption of linear MDP, we have $\nn{\phi(\cdot, \cdot)} \leq 1$ and $r_h(\cdot, \cdot) \leq 1$, so the first summand in both terms is at most $1$. Furthermore, $V_{h+1}^i(x_{h+1}^i)$ and $V_{h+1}^i({x'}_{h+1}^i)$ are at most $H$. Using the assumption that  $\nn{\phi(\cdot, \cdot)} \leq 1$, the second summand in each term is at most $H$. Combining these bounds, we arrive at the claim.

By definition of neighboring sequences and triangle inequality, the sensitivity of $\LambdaMAT_h^k$ can be written as the sum of $\nn{\phi(x_h^k, a_h^k)\phi(x_h^k, a_h^k)^\top}$ and $\nn{\phi({x'}_h^k, {a'}_h^k)\phi({x'}_h^k, {a'}_h^k)^\top}$. Since $\nn{\phi(\cdot, \cdot)} \leq 1$, the sensitivity of $\LambdaMAT_h^k$ is at most $2$.
\end{proof}
We release these two types of statistics with two mechanisms. Next, we show that both the Binary mechanism and the Gaussian mechanism each satisfies $\rho/2$-zCDP.

\paragraph{Analysis of Binary Mechanism:}
 First, we use the Gaussian binary mechanism \citep{shariff2018differentially} (a variant of the binary mechanism due to \cite{chan2011private, panprivacy} that preserves positive definiteness (PD) in matrices) to privatize the statistics $\{\LambdaMAT_h^k\}_{k,h\in[K]\times[H]}$. Here we show that releasing the sequence of private Gram matrices, denoted by $\{\Lambtil_h^k\}_{k,h\in[K]\times[H]}$,  satisfies $\rho/2$-zCDP and also, with high probability, the all are PD matrices.

The algorithm initializes $H$ private counters $\PC$ for $h\in[H]$ such that each satisfies $(\rho/(2H))$-zCDP. Denote $\phi_h^k = \phib{x_h^k, a_h^k}$ as the data observed during episode $k$.
Each counter $\PC$ observes a stream of Gram matrices  
$U_h = (\phi_h^1{\phi_h^1}^\top, \ldots, \phi_h^K{\phi_h^K}^\top)$
over $K$ episodes
and maintains a binary tree in which each internal node represents a partial sum. 
Let $\Sigma^{i:j}$ be the function of the input stream $U_h$ that computes the partial sum of all events between between time $i$ and $j$, defined as $\Sigma^{i:j}(U_h) = \sum_{\tau=i}^{j-1} \phi_h^\tau(\phi_h^\tau)^\top$.
The sum of all events before episode $k$ is can be computed by a function $\Sigma^{1:k}$, which is the output of at most  $\log(K)$ partial sums. Note that $\Delta(\Sigma^{1:k})\leq \log(K)$.
On each episode $k$, the private counter for $h$ outputs a random matrix $\PCkth{k}$ using the Gaussian mechanism as follows:
%
\begin{equation}
\label{eq:bincounter}
\PCkth{k} \leftarrow \Gau\pp{U_h, \Sigma^{1:k}, \frac{\log(K)^2}{2\rho}}
\end{equation}
Each partial sum satisfies $\pp{\frac{\rho}{ 2H\log(K)}}$-zCDP. 
However, since any single data point in $U_h$ appears in at most $\log(K)$ nodes of $\PC$, we must do  composition over $\log(K)$ events. It follows that the output of $\PC$ satisfies $\frac{\rho}{2H}$-zCDP. By another round of composition over all time steps $H$, we have that releasing $K\cdot H$ counts given by $\{\PCkth{k}\}_{k\in [K], h\in [H]}$ satisfies $\rho/2$-zCDP.

The noise added by the \BM can violate the requirement that
$\Lambtil_h^k$ is positive definite (PD). For that reason, in line \ref{lin:shiftgram},  we shift the noisy Gram matrix of $\PC$ by a constant $2\binmechboundvar\I$.

Next we show that the shifted noisy Gram matrix $\Lambtil_h^k$ is PD, by showing that all eigenvalues are strictly positive.  
First, let $\M_h^k$
be the noise added by the \BM~ on episode $k$ such that $\Lambtil_h^k = \sum_{\tau=1}^{k-1} \phi_h^\tau(\phi_h^\tau)^\top + 2\binmechboundvar\I + \M_h^k$, then we only need to show that $2\binmechboundvar\I + \M_h^k$ is PD. By known concentration bounds \citep{tao2012topics} on the matrix operator norm state that with probability at least $1-p$  on all for $k, h\in[K]\times [H]$:
\begin{align}\label{eq:operatornormbound}
\| \M_h^k \|_{op}    \leq \binmechboundvar \coloneqq \binmechbound
\end{align}
If the eigenvalues of $\M_h^k $ are $\lambda_1,\ldots, \lambda_d$, then $\M_h^k + 2\binmechboundvar\I$ has eigenvalues  $\lambda_1 + 2\binmechboundvar,\ldots, \lambda_d + 2\binmechboundvar$. 
 By definition of operator norm and  \cref{eq:operatornormbound} we have $\max_{i\in[d]} |\lambda_i| =  \|\M_h^k\|_{op} \leq \binmechboundvar$.
Therefore, all eigenvalues of $\Lambtil_h^k $ are positive and thus the matrix is PD.

\paragraph{Analysis of Gaussian Mechanism:}
For the remaining of this section, the focus is to show that releasing the statistics $\{\ytil_h^{\tilde{k}}\}_{k,h\in[K]\times[H]}$
also satisfies $(\rho/2)$-zCDP. Recall that, if $k$ is an update episode then \cref{alg:privrl} adds Gaussian noise to the statistics $\y_h^k$. 
Naively, we could update the episode every round, but the noise added would scale with $O(\rho / K)$, giving an error in the order of $O(\sqrt{K/(\rho )})$, which leads to sub-optimal regret on the variable $K$. Therefore, one would want to update the policy at most $O(dH\log(K))$ times because it allows us to scale the noise with $\rho/\log(K)$, which gives logarithmic error on the parameter $K$. 

The main challenge is that the update episode  is a random variable dependent on the user sequence. Thus, any single user in the sequence affects the future update episodes. Therefore, to simplify the analysis, we pretend to run a hypothetical Gaussian mechanism to release a constant function during rounds in which an update is not triggered. Formally, for any $h\in[H]$, we construct an adversary that selects an adaptive sequence of functions $f_h^1,\ldots, f_h^K$ as follows: On episode $k$, if $k$ is an update episode then the adversary sets $f_h^k(U_k) = \sum_{i=1}^{k-1} \phib{x_h^i,a_h^i} r_h^i\defeq \y_h^k$ for all $h\in[H]$. Otherwise, the adversary sets $f_h^k(U_k)=0$ for all $h\in[H]$. On non-update episodes, $f_h^k$ is a constant function independent of the data. Also,  \cref{alg:privrl} uses $f_h^k$ only on update episodes to modify the policy and ignores it otherwise. 

The following lemma gives a tool to analyze the protocol under zCDP as long as the joint sensitivity of the sequences $\{f_h^1,\ldots, f_h^K\}_{h\in[K]}$ is bounded.

\begin{lemma}\label{lem:gaussianadaptivecomp}
For any $K>0$,  let $\{f_k:\cU\rightarrow \mathbb{R}^d\}_{k\in[K]}$ be a sequence of adaptively chosen functions, such that the joint sensitivity is bounded by: 
$\sum_{k=1}^K \Delta(f_i) \leq  \Delta$. Then, the composition of $\{f_k(D_k^b) + \eta_k\}_{k\in[K]}$,   where $\eta_k\sim \cN(0, \frac{\Delta}{\rho})$  satisfies $\rho$-zCDP.
\end{lemma}


It only remains to show that the joint sensitivity of the adaptive sequences is bounded. 
First, we show that each $f_h^k$ has bounded sensitivity. By definition of $f_h^k$ and \cref{lem:sensitivity}, we have on any update episode its sensitivity is given by $\Delta(f_h^k) = \Delta(\y_h^k) \leq 3H$  , and on non-update episodes we have that the sensitivity is $\Delta(f_h^k)=0$ for any $h\in [H]$.

Next, we use the fact that number of update episodes is bounded by $\switchcostvar \defeq \switchingcost$.
Then the joint sensitivity of the the adaptive sequences is given by
\begin{align*}
   \sum_{h=1}^H\sum_{k=1}^K
   \Delta(f_h^k) 
   = 
   \sum_{h=1}^H\sum_{k=1}^K
   \Delta(\y_h^k) 
   \leq 3H^2\switchcostvar
\end{align*}


Therefore releasing $f_h^k$ with a $\pp{(1/2)\rho / \pp{3H^2\switchcostvar}}$-zCDP Gaussian mechanism on every episode $k\in[K]$ satisfies $\rho/2$-zCDP. This concludes the proof.

\subsection{Regret Analysis}\label{sec:regret}

After showing that \Cref{alg:privrl} is JDP, we finish the analysis by showing the regret guarantees in the following theorem. 

\begin{theorem}[Regret]\label{thm:regret}
For any $p\in (0,1)$, any privacy parameter $\rho>0$, if we set  $C = 2$ and 
$$\beta=\betabound$$
with $\chi = \chibound$ in \cref{alg:privrl}, then with probability $1-p$ , the total regret of Private LSVI-UCB (\cref{alg:privrl}) is at most 
\Cref{alg:privrl} satisfies $(\epsilon, \delta)$-differential privacy with $\epsilon = \rho + 2 \sqrt{\rho \log(1/\delta)}$. If $\epsilon < 1$, then the total regret of \Cref{alg:privrl} is at most
\begin{align*}
    R(K) \leq \widetilde{O} \left ( d^{3/2} H^2 K^{1/2} + \frac{H^3 d^{5/4} K^{1/2} \pp{\log \left(\tfrac{1}{\delta} \right)}^2}{\epsilon^{1/2}}  \right)
\end{align*}
\end{theorem}

\begin{proof}[Proof Sketch]
We present a proof sketch here for the regret
analysis. Here the $\widetilde{O}(\cdot)$ notation hides all $\log$ and constants terms. See Appendix B for full detail of the proof. The analysis is similar to \citet{jin2020provably}, but there are several challenges we need to handle
due to privacy noise. The proof structure is as follows: (1) Show that the privacy cost to preserve the set of sufficient  statistics scales with $O(\log(K)$. (2) Construct new confidence bounds given the private statistics and show that the action-value function from \cref{alg:privrl} is optimistic. (3) As in \citet{jin2020provably} and \citet{wang2021provably}, we follow an optimism strategy and upper bound the regret by the sum of bonus terms: $\sum_{k=1}^k\sum_{h=1}^H\text{B}_h^k(x_h^k, a_h^k)$, where $B$ is defined in line \ref{lin:bonus} of \Cref{alg:privrl}.
%



\paragraph{Private Least Squares}
As described in \cref{sec:algorithm}, \cref{alg:privrl} uses a Least-Squares-Value-Iteration (LSVI) to estimate the action-value function $Q_h^{\star} \defeq \vdot{\phi(x,a), \w_h^{\star}}$, but to preserve privacy, \cref{alg:privrl} adds noise to the sufficient statistics. For any $k,h \in [K] \times [H]$, let $\LambdaMAT_h^k$ be the gram matrix as defined in \cref{eq:lambdadef} and $\y_h^k$ be as defined in \cref{eq:ydef}. Let their privatized statistics be
%
\begin{align}
    \Lambtil_h^{k} = {\LambdaMAT_h^k} + 2\binmechboundvar\I + \M_h^k, \quad \quad  
    \ytil_h^k = {\y_h^k} + \etabf_h^k
\end{align}
where  $\{\M_h^k, \etabf_h^k\}_{k,h\in[K]\times [H]}$ are noise injected to privatize the statistics. 
Let $\lambda_{\text{min}}(\cdot)$ denote the minimum eigenvalue of a matrix, then by \cref{eq:operatornormbound}  we have a lower bound on the minimum eigenvalue of $\Lambtil_h^k$:
\begin{align}
    \lambda_{\text{min}}(\Lambtil_h^k) \geq \binmechboundvar \coloneqq \binmechbound
\end{align}

Also, by the tail bound on the Gaussian noise we have $\|\etabf_h^k\|_2 \leq \widetilde{O}\pp{H d /\sqrt{\rho}}$. It follows that 
\begin{align}
    \label{eq:etabound}
    \|\etabf_h^k\|_{(\Lambtil_h^k)^{-1}}  
    \leq \tfrac{1}{\sqrt{\binmechboundvar}}\|\etabf_h^k\|_2 
    %
    %
    \leq \widetilde{O}\pp{H^2 \sqrt{d \binmechboundvar}}
\end{align}


%
Recall that  LSVI, on episode $k$, starts with $h=H$ and recursively computes $\w_h^{k}= (\Lambtil_h^k)^{-1} \ytil_h^k $. 
%
\paragraph{Upper Confidence Bounds (UCB)} 
Let $\pi$ be any policy with corresponding action-value $Q_h^\pi(x,a)=\vdot{\phi(x,a), \w_h^{\pi}}$. Let $\widehat{Q}_h^k(x,a) = \vdot{\phi(x,a), \w_h^{k}}$ be the private empirical action-value function induced by $\w_h^k$ without the bonus term. We give a confidence bound around $\widehat{Q}_h^k$ 
and construct an optimistic action-value function $Q_h^k$ (i.e., ${Q}_h^k(x,a)\geq Q_h^\star(x,a)$ for all $(x,a)$), taking into account  the noise added to privatize the statistics. 
We begin by decomposing the term $ \w_h^{k} - \w_h^\pi$ as follows:
\begin{align*}
    \w^{k}_h -  \w^{\pi}_h
     =\underbrace{ \pp{\Lambtil_h^{k}}^{-1} \pp{ \y_h^k} - \w^{\pi}_h}_{\va}
     + \underbrace{\pp{\Lambtil_h^{k}}^{-1} \etabf_h^{k}}_{\vb} 
\end{align*}
Then, for all $(x,a, h,k)$,  we can write
$$\widehat{Q}_h^k(x,a)-{Q}_h^\pi(x,a)= \vdot{\phi(x,a), \va + \vb}$$ and bound each term  independently.
Next we directly use the Lemma  B.4 in \citet{jin2020provably} to   bound $\vdot{\phi(x,a), \va}$  by the  expected difference at next step, plus an error term that depends on the minimum eigenvalue of the gram matrix, which in our case is $\binmechboundvar$. Therefore, we obtain:
\begin{align}
    \left| \vdot{\phi(x,a), \va}     \right|  
    &\leq {\mathbb{P}_h(V_h^k - V_h^{\pi})(x,a)} + 
    \widetilde{O}\pp{
    H\sqrt{d \binmechboundvar } + dH
    }\|\phi(x,a) \|_{(\Lambtil_h^{k})^{-1}}
    \label{eq:A}
\end{align}
where $\mathbb{P}_h(V_h^k - V_h^{\pi})(x,a)$ is the expected error  in time step $h+1$ between $V_{h+1}^k$ and $V_{h+1}^\pi$ after taking action $(x,a)$.
It remains to give a confidence bound for the second term $\vdot{\phi(x,a), \vb}$, which follows simply by the Cauchy-Schwartz inequality:
\begin{align}
    \label{eq:B}
|\vdot{\phi(x,a), \vb}|
    \leq  \nn{\etabf_h^{k}}_{(\Lambtil_h^{k})^{-1}}
    \cdot 
    \nn{\phi(x,a)}_{(\Lambtil_h^{k})^{-1}}
\end{align}
Note that by the bound in \eqref{eq:etabound}, the term $\nn{\etabf_h^{k}}_{(\Lambtil_h^{k})^{-1}}$ is smaller than the second term in \eqref{eq:A}. 
Therefore, choosing $\beta=\widetilde{O}\pp{H^2 \sqrt{d \binmechboundvar} + dH}$ is enough to 
upper bound both \eqref{eq:A} and \eqref{eq:B}.  That is, with high probability, for any policy $\pi$ and all $(x,a,k,h)$ we have: 
\begin{align}
\label{eq:Qconfidenceinterval}
    \widehat{Q}_h^k(x,a)-{Q}_h^\pi(x,a) \leq 
    \mathbb{P}_h(V_h^k - V_h^{\pi})(x,a)
    +
    \beta\|\phi(x,a) \|_{(\Lambtil_h^{k})^{-1}}
\end{align}
Furthermore, by the upper bound \eqref{eq:Qconfidenceinterval} 
the  optimistic action-value function is given by  $Q_h^k(x,a) = \vdot {\phi(x,a), \w_h^k} + \beta \|\phi(x,a) \|_{(\Lambtil_h^k)^{-1}}$. 
That is for all $(x,a,k,h)$:
\begin{align}\label{eq:Qoptimism}
    Q_h^k(x,a)  \geq  Q_h^\star(x,a)
\end{align}
\paragraph{Regret proof} 
In this section we use the following notation $\phi_h^k = \phi\pp{x_h^k, a_j^k}$. Note that on episode $k$, the agent acts according to policy $\pi_k$ which is given by  $\pi_{k}(x,h) = \argmax_{a} Q_h^{\tilde{k}}(x,a)$, where  $\tilde{k} \leq k$ be the
last episode the algorithm updated its policy for time step $h$.

By optimism \eqref{eq:Qoptimism}, the regret can be upper bounded by the difference of the optimistic 
value function  $V_h^k(\cdot) = \max_{a} Q_h^k(\cdot, a)$ and the value function induced by policy $\pi_k$:
\begin{align*}
    R(K) = \sum_{k=1}^K  V_1^\star(x_1^k) - V_1^{\pi_{k}}(x_1^k)
    \leq \sum_{k=1}^K  V_1^{\tilde{k}}(x_1^k) - V_1^{\pi_{k}}(x_1^k)
\end{align*}
For the next step we set $\delta_h^k = V_h^{\ktil}(x_h^k) -  V_h^{\pi_k}(x_h^k)$ and $\zeta_{h+1}^k = \mathbb{P}_h(V_h^\ktil - V_h^{\pi})(x_h^k,a_h^k) - \delta_{h+1}^k$.
Then by the bound in \eqref{eq:Qconfidenceinterval}, we can recursively relate the error on time step $h$ to the error on time step $h+1$. The results is:
\begin{align*}
    \delta_h^k \leq 
    \delta_{h+1}^k  +
     \underbrace{ \mathbb{P}_h(V_h^\ktil - V_h^{\pi})(x_h^k,a_h^k)
     - \delta_{h+1}^k}_{\zeta_{h+1}^k}
    + \beta \|\phi_h^k \|_{(\Lambtil_h^\ktil)^{-1}}
\end{align*}
Unfolding the recursion, we have that with high probability, the regret of \cref{alg:privrl} is upper bounded by:
\begin{align}\label{eq:regretdecomposition}
     R(K)&\leq 
     \underbrace{
        \sum_{k=1}^K \sum_{h=1}^H \tilde{\zeta}_h^k
     }_{\leq \widetilde{O}\pp{\sqrt{KH^3}}}
     +
     \underbrace{
     3 \beta \sum_{k=1}^K \sum_{h=1}^H 
     %
     \beta \|\phi_h^k \|_{(\Lambtil_h^\ktil)^{-1}}
     }_{
        \leq \widetilde{O}\pp{ H\beta \sqrt{d K } }
        }
\end{align}
The first term of \eqref{eq:regretdecomposition} is the sum of a zero-mean martingale difference sequence bounded by $2H$.  Thus, using standard concentration inequalities it's at most $\widetilde{O}\pp{2\sqrt{K H^3}}$ with high probability.
For the second term, the first step is to show that the low switching cost constraint only adds a constant to the regret. This can be seen by the following inequality:
The $\|\phi_h^k \|_{(\Lambtil_h^\ktil)^{-1}} \leq \sqrt{2}\|\phi_h^k \|_{(\Lambtil_h^k)^{-1}}$. The final bound of the second term in \eqref{eq:regretdecomposition} follows from 
an application of Cauchy-Schwartz inequality and the trace-determinant lemma and elliptical potential lemma from \citet{abbasi2011improved}.
The final bound on the second term is $\widetilde{O}\pp{ H\beta \sqrt{d K }}$. 
\end{proof}

\section{Conclusion} Our algorithm follows an adaptive policy update schedule and benefits from low switching cost, i.e only performs $O(\log(K))$ updates, compared to $O(\mathrm{poly}(K))$ updates in previous works using non-adaptive batching. However, the update time is an unknown random variable depending on the data and randomness of the algorithm, which introduces new challenges in privacy analysis. We view the interactions between the algorithm and the users as an adaptive adversary with bounded sensitivity. Hence, our privacy cost is only proportional to the number of updates, instead of the number of episodes. Our low-switching cost algorithm and analysis advances the existing paradigm using binary mechanism and provides a foundation for future works in private RL.

\section*{Acknowledgement}
ZSW, DN, GV were supported in part by the NSF FAI Award $\#1939606$, NSF SCC Award $\#1952085$, a Google Faculty Research Award, a J.P. Morgan Faculty Award, a Facebook Research Award, and a Mozilla Research Grant. Any opinions, findings, conclusions, or recommendations expressed in this material are those of the authors and not necessarily reflect the views of the National Science Foundation and other funding agencies. 
\bibliography{main}

\bibliographystyle{icml2022}

\newpage 
\appendix 


\section{Privacy Analysis}
\paragraph{Proof of \Cref{lem:sensitivity}}
Let $U$ and $U'$ be any two neighboring user sequences, and $\y_h^k, (\y')_h^k$ are the outputs of these two user sequences, respectively. Let $i \leq k$ be some episode with $x_h^i \neq {x'}_h^i$ and $a_h^i \neq {a'}_h^i$, where $(x_h^i, a_h^i) \in U$ and $({x'}_h^i, {a'}_h^i) \in U'$. Then, by definition we have:
\begin{align*}
     &\quad \nn{\y_h^k- (\y')_h^k}_2\\
     &= \Bigg\| \sum_{j=1}^{k-1} \phi(x_h^j, a_h^j)
    \left(
    r_h(x_h^j, a_h^j) + V_{h+1}^j(x_{h+1}^j)
    \right) - \sum_{j=1}^{k-1} \phi({x'}_h^j, {a'}_h^j)
    \left(
    r_h({x'}_h^j, {a'}_h^j) + V_{h+1}^j({x'}_{h+1}^j)
    \right) \Bigg\|_2\\
     &= \Bigg\| \phi(x_h^i, a_h^i) 
    \left(r_h(x_h^i, a_h^i) + V_{h+1}^i(x_{h+1}^i) \right) -
    \phi({x'}_h^i, {a'}_h^i)
    \left(r_h({x'}_h^i, {a'}_h^i) + V_{h+1}^i({x'}_{h+1}^i) \right) \Bigg\|_2 \tag{since the two user sequences only differs in the $i$-th user.} \\
    &\leq 
    2 + \left\|
    \phi(x_h^i, a_h^i) V_{h+1}^i(x_{h+1}^i)
    - 
    \phi({x'}_h^i, {a'}_h^i) V_{h+1}^i({x'}_{h+1}^i)\right\|_2 \tag{by linear MDP assumption, $r_h(x_h^i, a_h^i)\leq 1$ and 
$\| \phi(x_h^i, a_h^i)\|_2\leq 1$} \\
    &\leq 2 + \left \| \phi(x_h^i, a_h^i) V_{h+1}^i(x_{h+1}^i) \right \|_2 + \left \| \phi({x'}_h^i, {a'}_h^i) V_{h+1}^i({x'}_{h+1}^i) \right \|_2 \tag{by triangle inequality}\\
    &\leq 2 + \left | V_{h+1}^i(x_{h+1}^i) \right | + \left | V_{h+1}^i({x'}_{h+1}^i) \right | \tag{by linear MDP assumption, $\|\phi(.,.)\|_2 \leq 1$}\\ 
    &\leq 2 + 2H 
\end{align*}

Observe that for $\Delta(\LambdaMAT_h^k)$, by definition, we have

\begin{align*}
    &\quad\nn{\LambdaMAT_h^k - (\LambdaMAT')_h^k} \\
    &= \nn{\left( \lambda \I + \sum_{j=1}^{k-1} \phi(x_h^j, a_h^j) \phi(x_h^j, a_h^j)^\top \right) - \left( \lambda \I + \sum_{j=1}^{k-1} \phi({x'}_h^j, {a'}_h^j) \phi({x'}_h^j, {a'}_h^j)^\top \right)}_2\\
    &= \nn{ \phi(x_h^i, a_h^i) \phi(x_h^i, a_h^i)^\top - \phi({x'}_h^i, {a'}_h^i) \phi({x'}_h^i, {a'}_h^i)^\top}_2 \tag{since the two user sequences only differ in the $i$-th user}\\
    &\leq \nn{\phi(x_h^i, a_h^i) \phi(x_h^i, a_h^i)^\top}_2 + \nn{\phi({x'}_h^i, {a'}_h^i) \phi({x'}_h^i, {a'}_h^i)^\top}_2 \tag{by triangle inequality}\\
    &\leq 2 \tag{Note that $\nn{\phi(x_h^t, a_h^t)}_2 \leq 1$}
\end{align*}
With these two bounds, we conclude our proof.




\section{Full Regret Analysis}\label{sec:regretfull}


\subsection{Notation and math tools}\label{sec:notation}

We start this section by listing our commonly used notations:

\begin{multicols}{2}
\begin{itemize}
    \item $\lambda \defeq 1$ 
    \item $\lambtil_{y} \defeq \gaumechbound$  
    \item $\lambtil_{\Lambda} \defeq \binmechbound$  
    \item $U_K \defeq \max \left\{ 1, 2H \sqrt{\frac{dK}{ \binmechboundvar}} + \frac{\gaumechboundvar}{ \binmechboundvar} \right \}$
    \item $\beta \defeq \betabound$
    \item $ \M_h^t \defeq \Lambtil_h^{t} - \LambdaMAT_h^{t}, \quad \etabf_h^k \defeq \ytil_h^k - \y_h^k$
    \item For any episode $k\in[K]$, $\tilde{k} < k$ is the last update episode before $k$. 
    \item $\chi \coloneqq \chibound$.
\end{itemize}
\end{multicols}

\paragraph{Notation (as in \cite{jin2020provably})}

\begin{fact}
For any two PSD matrices $A,B\in \mathbb{R}^{d\times d}$.
If $B \succ A$ then for any vector $x\in\mathbb{R}^d$  we have $\|x\|_A \leq \|x\|_B $ and $\|x\|_{B^{-1}} \leq \|x\|_{B^{-1}}$.
\end{fact}
\begin{proof}
Since $A,B$ are PSD and $B\succ A$ then we have that $B-A\succ 0$ is also PSD. Then by definition of PSD we have $x^\top (B-A)x\geq 0$, which implies that $ x^\top (B)x \geq x^\top (A)x$.
\end{proof}

\subsection{Private statistics}


Recall from \ref{sec:notation} that  $ \M_h^k \defeq \Lambtil_h^{k} - \LambdaMAT_h^{k}$ and   $\etabf_h^k \defeq \ytil_h^k - \y_h^k$. Then consider the following events:
\begin{align}
\label{eq:privacyeventI}
    \mathcal{E}_1 &:= \left\{\forall {(k,h)\in [K]\times [H] :
   \left \| \etabf_{h}^k \right\|_2 \leq \lambtil_{y}}\right\}
   \text{, } \quad \lambtil_y = \gaumechbound
   \\ 
\label{eq:privacyeventII}
    \mathcal{E}_2 &:= \left\{\forall {(k,h)\in [K]\times [H] :
   \left\|\M_{h}^k \right\|_2 \leq \lambtil_{\Lambda}}\right\}\text{, } \quad \lambtil_\Lambda = \binmechbound\\
\label{eq:privacyeventIII}
    \mathcal{E}_3 &:= \left \{\forall(k,h)\in[K]\times [H] :
    \left\|
    \sum_{i=1}^{\tilde{k}-1 }\phi_h^i
    \left[
    V_{h+1}^{\tilde{k}} (x_{h+1}^i) 
    -
    \bbP_h V_{h+1}^{\tilde{k}} (x_h^i, a_h^i)
    \right]
    \right\|_{(\Lambtil_h^{\tilde{k}})^{-1}}
    \leq 6 d H \sqrt{\log (\chi)} \right \}
\end{align}
Consider event $\mathcal{E} = \mathcal{E}_1 \cup \mathcal{E}_2 \cup \mathcal{E}_3$.

\subsection{High probability events}
We will show that the probability of $\mathcal{E}$ is at most $p$. 
We begin by analyzing the error of the private algorithm.
The next theorem states the utility guarantees given by the privacy mechanism:
\begin{theorem}[Utility]\label{thm:utility}
Let $\{\Lambtil_h^{k}\}_{h\in[H], k\in[K]}$ and $\{\ytil_{h}^k\}_{h\in[H], k \in [K]}$ be the privatized sufficient statistics of \cref{alg:privrl} satisfying $\rho$-zCDP. 
Then with probability at least $1-p$ we have for any $k\in[K]$ and $h\in [H]$
\begin{equation*}
    \|\Lambtil_h^{k} - \LambdaMAT_h^{k} \|_{op}
    \leq \frac{16\sqrt{H}\log(K) \pp{4\sqrt{d+1} + 2\log\pp{\frac{KH}{p}}} }{\sqrt{2\rho}}
\end{equation*}
and for any $k\in[K]$ and $h\in [H]$ we have 
\begin{equation*}
    \|\ytil_h^k - \y_h^k \|_2 \leq \sqrt{ \frac{\pp{\switchingcost}^2\log\pp{\frac{KH}{p}}}{\rho}}
\end{equation*}
\end{theorem}
By \cref{thm:utility} we have that $\pr{ \mathcal{E}_1}\leq \frac{p}{3}$ and $\pr{ \mathcal{E}_2}\leq \frac{p}{3}$. From \Cref{lem:B3}, condition on event $\mathcal{E}_1 \cup \mathcal{E}_2$, we have $\Pr[\mathcal{E}_3] \leq \frac{p}{3}$. Then by union bound, $\Pr[\mathcal{E}] \leq p$.

\subsection{Lemmas from \citet{jin2020provably}}
Here we state key lemmas in the regret analysis. See \cref{sec:jinproofs} for the proofs.

\begin{lemma}[Lemma B.1 in \cite{jin2020provably}]
\label{lem:B1}
Under the linearity assumption, for any fixed policy $\pi$, let $\{\w^{\pi}_h \}_{h \in [H]}$ be the corresponding weights such that $Q^{\pi}_h(x,a) = \langle \phi(x,a), \w_h^{\pi} \rangle$ for all tuples $(x,a,h) \in \mathcal{S} \times \mathcal{A} \times [H]$. Then, we have 
\begin{equation*}
    \forall h\in [H], \nn{\w_h^{\pi}} \leq 2H\sqrt{d} 
\end{equation*}
\end{lemma}

\begin{lemma}[Lemma B.2 in \cite{jin2020provably}] Under the event $\mathcal{E}$, for any $(k,h) \in [K] \times [H]$, the parameter vector $\w_h^k$ in \Cref{alg:privrl} satisfies:
\begin{equation*}
    \nn{\w_h^k} \leq 
    U_K := \max \left\{ 1, 2H \sqrt{\frac{dK}{ \binmechboundvar}} + \frac{\gaumechboundvar}{ \binmechboundvar} \right \}
\end{equation*}
\label{lem:B2}
\end{lemma}

\begin{theorem}[Theorem D.3 in \cite{jin2020provably}]
Let $\{ \epsilon_i \}_{i=1}^\infty$ be a real-valued stochastic process with corresponding filtration $\{ \mathcal{F}_i \}^\infty_{i=0}$. Let $\epsilon_i|\mathcal{F}_{i=1}$ be zero-mean and $\sigma$-subGaussian; i.e. $\E[\epsilon_i| \mathcal{F}_{i-1}] = 0$, and 
\begin{equation*}
    \forall \lambda \in \mathbb{R}, \quad \E[\exp(\lambda \epsilon_i) | \mathcal{F}_{i-1}] \leq \exp(\lambda^2 \sigma^2 /2 )
\end{equation*}
Let $\{\phi_i \}_{i=0}^\infty$ be an $\mathbb{R}^d$-valued stochastic process where $\phi_i \in \mathcal{F}_{t-1}$. Assume $\Lambda_0$ is a $d \times d$ positive definite matrix, and let $\Lambtil_i = \Lambtil_0 + \sum_{s=1}^{i} \phi_s \phi_s^\top + \M_i$. Then for any $\delta > 0$, with probability at least $1 - \delta$, we have for all $i \geq 0$: 
\begin{equation*}
    \nn{\sum_{s=1}^i \phi_s \epsilon_s}^2_{(\Lambtil^i)^{-1}} \leq 2 \sigma^2 \log \left[ \frac{\det(\Lambtil_i)^{1/2} \det(\Lambtil_0)^{-1/2}}{\delta} \right] 
\end{equation*}
\label{thm:D3}
\end{theorem}

\begin{lemma}[Lemma D.4 in \cite{jin2020provably}] Let $\{x_i\}_{i=1}^{\infty}$ be a stochastic process on state space $\mathcal{S}$ with corresponding filtration $\{ \mathcal{F}\}_{i=0}^\infty$. Let $\{ \phi_i\}_{i=0}^\infty$ be an $\mathbb{R}^d$-valued stochastic process where $\phi_i \in \mathcal{F}_{i-1}$, and $\nn{\phi_i} \leq 1$. Let $\Lambtil = 2\binmechboundvar \I+ \M + \sum_{i=1}^{\tilde{k}} \phi_i \phi_i^\top  $. Then for any $\delta > 0$, with probability at least $1 - \delta$, for all $k \geq 0$, and any $V \in \mathcal{V}$ so that $\sup_{x} |V(x)| \leq H$, we have:
\begin{equation*}
    \nn{\sum_{i=1}^k \phi_i \{V(x_i) - \E[V(x_i)| \mathcal{F}_{i-1}] \}}^2_{(\Lambtil^{\tilde{k}})^{-1}} \leq 4H^2 \left[ \frac{d}{2} \log \left( \frac{k + \lambtil_\Lambda}{ \lambtil_\Lambda} + \log \left( \frac{\mathcal{N}_{\epsilon}}{\delta} \right) \right) \right] + \frac{8k^2 \epsilon^2}{ \lambtil_\Lambda}
\end{equation*}
where $\mathcal{N}_\epsilon$ is the $\epsilon$-covering of $\mathcal{V}$ with respect to the distance $\mathrm{dist}(V,V') = \sup_x | V(x) - V'(x)|$.
\label{lem:D4}
\end{lemma}

\begin{lemma}[Lemma D.5 in \cite{jin2020provably}]
For any $\epsilon > 0$, the $\epsilon$-covering number of the Euclidean ball in $\mathbb{R}^d$ with radius $R > 0$ is upper bounded by $(1 + 2R/\epsilon)^d$.
\label{lem:D5}
\end{lemma}

\begin{lemma}[Similar to Lemma D.6 in \cite{jin2020provably}] \label{lem:D6}
Let $\mathcal{V}$ denote a class of functions mappings from $\mathcal{S}$ to $\mathbb{R}$ with following parametric form:
\begin{align*}
    V(.) = \min \left \{ \max_a \w^\top \phi(., a) + \beta \sqrt{\phi(.,a)^\top (\Lambtil)^{-1} \phi(.,a)}, H \right \}
\end{align*}
where the parameters $(\w, \beta, \Lambtil)$ satisfy $\nn{\w} \leq L, \beta \in [0, B]$ and the minimum eigenvalue satisfies $\lambda_{\min}(\Lambtil) \geq \lambtil_\Lambda$. Assume $\nn{\phi(x,a)} \leq 1$ for all $(x,a)$ pairs, and let $\mathcal{N}_{\epsilon}$ be the $\epsilon$-covering number of $\mathcal{V}$ with respect to the distance $\mathrm{dist}(V, V') = \sup_x | V(x) - V'(x)|$. Then,
\begin{equation*}
    \log(\mathcal{N}_\epsilon) \leq d \log(1 + 4L/\epsilon) + d^2 \log( 1 + 8 d^{1/2} B^2 / ( \lambtil_\Lambda\epsilon)^2) 
\end{equation*}
\end{lemma}

\begin{lemma}[Similar to Lemma B.3 in \cite{jin2020provably}]
\label{lem:B3}
Under the event $\mathcal{E}_1 \cup \mathcal{E}_2$, 
for $\chi = \chibound$ and $\beta =\betabound$,
for any fixed $p\in(0,1)$ if we let  $\mathcal{E}_3$ be the event that
\begin{align*}
    \forall(k,h)\in[K]\times [H] :
    \left\|
    \sum_{i=1}^{\tilde{k}-1 }\phi_h^i
    \left[
    V_{h+1}^{\tilde{k}} (x_{h+1}^i) 
    -
    \bbP_h V_{h+1}^{\tilde{k}} (x_h^i, a_h^i)
    \right]
    \right\|_{(\Lambtil_h^{\tilde{k}})^{-1}}
    \leq 6 d H \sqrt{\log (\chi)}
\end{align*}
 
Then $\bbP(\mathfrak{E}| \mathcal{E}) \geq 1 - \frac{p}{3}$
\end{lemma}

Recall that our value function for episode $k$ and timestep $h$ is given by $Q_h^t(x,a) = \vdot{\phi(x,a), \w_h^{\tilde{k}}}$, where $\tilde{k}<k$ is the last update episode.

\begin{lemma}[Similar to Lemma B.4 in \cite{jin2020provably}] 
\label{lem:B4}
Under event $\mathcal{E}_1 \cup \mathcal{E}_2$, 
for any fixed policy $\pi$, on the event $\mathcal{E}_3$ defined in \cref{lem:B3}, we have for all $(x,a,h,k)$ that:
\begin{align*}
    Q_h^{\tilde{k}}(x,a) - Q_h^\pi(x,a) 
    =\bbP_h\pp{{\V}_{h+1}^{\tilde{k}} - \V_{h+1}^\pi}(x,a) + \Delta_h^k(x,a)
\end{align*}
for some $\Delta_h^k(x,a)$ that satisfies $|\Delta_h^k(x,a)| \leq \beta \nn{\phi(x,a)}_{ (\Lambtil_h^{\tilde{k}})^{-1}}$.
\end{lemma}

\begin{lemma}[Lemma B.5 in \cite{jin2020provably}, (UCB)] \label{lem:B5}
Under the event $\mathcal{E}$, we have that $Q_h^{\tilde{k}}(x,a) \geq Q_h^*(x,a)$ for all $(x,a,h,k) \in \mathcal{S} \times \mathcal{A} \times [H] \times [K]$.

\end{lemma}

\begin{lemma}[Lemma B.6 in \cite{jin2020provably}(Recursive Lemma)]
\label{lem:B6}
Let $\tilde{k}$ be the last update episode before $k$.
For any $(h,k) \in [H] \times [K]$, let $\tilde{\delta}_h^k = {V}_h^{\tilde{k}}(x_h^k) - V_h^{\pi_{\tilde{k}}}(x_h^k)$ denotes the errors of the estimated ${V}_h^{\tilde{k}}$ relative to $V_h^{\pi_{\tilde{k}}}$. 
Let $\zeta_{h+1}^k = \mathbb{E}[\tilde{\delta}_{h+1}^k | x_h^k, a_h^k] - \tilde{\delta}_{h+1}^k$. Then condition on the event $\mathfrak{E}$ defined in $\Cref{lem:B3}$, we have the following: for any $(k,h) \in [K] \times [H]$
\begin{equation*}
    \tilde{\delta}_h^k \leq \tilde{\delta}_{h+1}^k + \tilde{\zeta}_{h+1}^k + 2 \beta \nn{(\phi_h^{\tilde{k}})}_{ (\Lambtil_h^{\tilde{k}})^{-1}}  
\end{equation*}
\end{lemma}

\subsection{Final Regret Proof}

Finally, we prove the regret bound stated in \Cref{thm:regret}. 
For any $k \in [K]$, let $\tilde{k} \leq k$ be the
last episode the algorithm updated its policy.
That is, on episode $k$ the agent acts according to policy $\pi_k$ which is given by  $\pi_{k}(x,h) = \argmax_{a} Q_h^{\tilde{k}}(x,a)$.

We begin by decomposing the regret using the optimistic approximation. Thus, by \cref{lem:B5} we can decompose the regret as follows: 
\begin{align*}
        R(K) &= \sum_{k=1}^K [V_1^*(x_1^k) - V_1^{\pi_k}(x_1^k)]
            \leq \sum_{k=1}^K [{V}_1^{\tilde{k}}(x_1^k) - V_1^{\pi_{k}}(x_1^k)]
    = \sum_{k=1}^K \tilde{\delta}_1^k
\end{align*}

Recall that $ \tilde{\delta}_h^k = {V}_h^{\tilde{k}}(x_h^k) - V_h^{\pi_{k}}(x_h^k)$ and $\zeta_{h+1}^k = \mathbb{E}[\tilde{\delta}_{h+1}^k | x_h^k, a_h^k] - \tilde{\delta}_{h+1}^k$.
On the event  $\mathfrak{E}$, by the recursive lemma \cref{lem:B6} we obtain the following: 

\begin{align*}
     R(K) &\leq \sum_{k=1}^K \sum_{h=1}^H \tilde{\zeta}_h^k + 3 \beta \sum_{k=1}^K \sum_{h=1}^H \sqrt{(\phi_h^k)^\top \pp{\Lambtil_h^{\tilde{k}}}^{-1} \phi_h^k}
\end{align*}

For the first term, since the computation of $V_h^{\tilde{k}}$ is independent of the new observation $x_h^k$ at episode $k$, we obtain that $\{ \zeta_h^k \}$ is a martingale difference sequence satisfying $\zeta_h^k\leq 2H$ for all $k,h\in[K]\times[H]$.
Therefore, we can apply the Azuma-Hoeffding inequality, for any $t>0$, we have :
\begin{align*}
    \pr{ \sum_{k=1}^K \sum_{h=1}^H \tilde{\zeta}_h^k > t}
    \leq 
    \exp\pp{\frac{-t^2}{2K H^3}}
\end{align*}

Hence, with probability at least $1 - p/4$, we have
\begin{align}\label{eq:azuma}
    \sum_{k=1}^K \sum_{h=1}^H \tilde{\zeta}_h^k \leq 
    \sqrt{2KH^3 \log(4/p)}
    \leq \sqrt{2KH^3 \log(\chi)}
\end{align}
where $\chi \defeq  \frac{24^2 \cdot 18 \cdot K^2 d U_K H  }{p}$.
On the event $\mathcal{E}_2$, the condition of \cref{thm:switchingcost} is satisfied, which means that the total number of updates is bounded by $\switchingcost$ and the condition $\switchcountvar > \switchcostvar$ from Line~\ref{lin:update} in \cref{alg:privrl}  never happens. It follows that on $\mathcal{E}_2$, the following bound holds: .
\begin{align}\label{eq:doubling}
    (\phi_h^k)^\top\pp{\Lambtil_h^{\tilde{k}}}^{-1} \phi_h^k \leq 2 (\phi_h^k)^\top \pp{\Lambtil_h^{k}}^{-1} \phi_h^k \text{  for all } (k,h)\in [K]\times [H]
\end{align}

Thus, by \cref{eq:azuma} and \cref{eq:doubling} we have
\begin{align*}
     R(K)
     &\leq 
    \sqrt{2KH^3 \log(\chi)}
     + 6 \beta \sum_{k=1}^K \sum_{h=1}^H \sqrt{(\phi_h^k)^\top \pp{\Lambtil_h^{k}}^{-1} \phi_h^k} \\
\end{align*}

The next step is to apply Cauchy-Schwarz inequality to the second term:

\begin{align*}
     R(K)
     &\leq  \sqrt{2KH^3 \log(\chi)}
     + 6 \beta \sum_{h=1}^H \sqrt{K}
     \sqrt{\sum_{k=1}^K  (\phi_h^k)^\top (\Lambtil_h^{k})^{-1} \phi_h^k}
\end{align*}

Now recall that $\Lambtil_h^{k} = \LambdaMAT_h^{k} + 2\binmechboundvar\I + \M_h^k$. We use the fact that on event $\mathcal{E}_2$ (\cref{eq:privacyeventII}), we have $\|\M_h^k\| \leq \binmechboundvar$ and thus $\M_h^k +  \binmechboundvar\I \succ 0$, which implies that the minimum eigenvalue of $\Lambtil_h^k$ is $\binmechboundvar$.
%
%
%
Then by  \Cref{thm:D3} from \cite{jin2020provably} and $\|\LambdaMAT_h^K\|\leq \binmechboundvar + K$ we have that

\begin{align}\label{eq:selfnorm}
    \sum_{k=1}^K  (\phi_h^k)^\top \pp{\Lambtil_h^{k}}^{-1} \phi_h^k
    \leq 2\log\pp{\frac{\det{\Lambtil_h^{K+1}}}{\det\Lambtil_h^1}}
    \leq 2d\log\pp{\frac{\binmechboundvar + K}{\binmechboundvar}}
    \leq 2d\log\pp{\chi}
\end{align}

%
%
Finally, by \cref{eq:selfnorm}, and by substituting $\beta = \betabound$, we can bound the total regret as:
\begin{align*}
    &R(K)  \\
    &\leq  \sqrt{2KH^3 \log(\chi)}
     + 6 \beta H
     \sqrt{2 d K \log\pp{\chi}} \\
     &\leq  \sqrt{2KH^3 \log(\chi)}
     +
     36 H^2 d^{3/2} \sqrt{2K\log(\chi)^2}
     + 30 H^3 d
     \sqrt{2 K \binmechboundvar \log\pp{\chi}^2} 
     %
\end{align*}

Recall that $\binmechboundvar = \binmechbound$. Therefore, by a union bound on the events $\mathfrak{E}, \mathcal{E}_1, \mathcal{E}_2$, and the event from \cref{eq:azuma} we have that with probability at least $1-p$,  the regret of \cref{alg:privrl} is  bounded by
\begin{align*}
    R(K) = \widetilde{O} \left ( d^{3/2} H^2 K^{1/2} + H^3 d^{5/4} K^{1/2} \rho^{-1/4} \right) 
\end{align*}

\subsection{Lemma proofs from \citet{jin2020provably}}\label{sec:jinproofs}

\paragraph{Proof of \cref{lem:B2} (Lemma B.2 in \citet{jin2020provably})}
For any unit vector $v \in \bbR^d$, we have:
\begin{align*}
    |v^\top \w_h^k| 
    &= \left| v^\top \pp{\Lambtil_h^{\tilde{k}}}^{-1}\ytil_k^{\tilde{k}} \right|\ \\
    &= \left|
    v^\top \pp{\Lambtil_h^{\tilde{k}}}^{-1}
    \pp{
    \sum_{i=1}^{\tilde{k}-1} \phi_h^i [r(x_h^i, a_h^i) + \max_a Q_{h+1}(x_{h+1}^i, a)] + \etabf_h^{\tilde{k}} 
    }
    \right|
    \\
    &\leq \sum_{i=1}^{\tilde{k}-1} 
    \left| v^\top \pp{\Lambtil_h^{\tilde{k}}}^{-1} \phi_h^i
    \right| (H + 1 ) 
    + \left| v^\top  \pp{\Lambtil_h^{\tilde{k}}}^{-1}  \etabf_h^{\tilde{k}} \right| \\
    &\leq \sqrt{\left [ \sum_{i=1}^{\tilde{k}-1} v^\top \pp{\Lambtil_h^{\tilde{k}}}^{-1} v  \right ] \left [ \sum_{i=1}^{\tilde{k} - 1} (\phi_h^i)^\top \pp{\Lambtil_h^{\tilde{k}}}^{-1} \phi_h^i \right ]} 2H + \nn{v} \nn{\pp{\Lambtil_h^k}^{-1}} \nn{\etabf_h^{\tilde{k}}}\\
    &\leq 2 H \sqrt{\frac{d(\tilde{k} - 1)}{\binmechboundvar}} + \frac{\lambtil_{y}}{ \binmechboundvar}\\
    &\leq 2 H \sqrt{\frac{dK}{\binmechboundvar}} + \frac{\lambtil_{y}}{ \binmechboundvar} \leq U_K
\end{align*}


\paragraph{Proof of \cref{lem:D4} (Lemma D.4 in \citet{jin2020provably})}

For any $V \in \mathcal{V}$, we know there exists a $\tilde{V}$ in the $\epsilon$-covering such that 
\begin{equation*}
    V = \tilde{V} + \Delta_V \quad\text{ and } \quad \sup_x |\Delta_V(x)| \leq \epsilon 
\end{equation*}
This gives the following decomposition:
\begin{align*}
    &\quad \nn{\sum_{i=1}^{\tilde{k}} \phi_i \{ V(x_i) - \E[V(x_i)| \mathcal{F}_{i-1}] \}}^2_{(\Lambtil^{\tilde{k}})^{-1}} \\
    &\leq 2 \nn{\sum_{i=1}^{\tilde{k}} \phi_i \{ \tilde{V}(x_i) - \E[\tilde{V}(x_i) | \mathcal{F}_{i-1}] \} }^2_{(\Lambtil^{\tilde{k}})^{-1}} + 2 \nn{\sum_{i=1}^{\tilde{k}} \phi_i \{ \Delta_V(x_i) - \E[\Delta_V(x_i) | \mathcal{F}_{i-1}] \} }^2_{(\Lambtil^{\tilde{k}})^{-1}} 
\end{align*}
where we can apply \Cref{thm:D3} and a union bound to the first term, and bound the second term by $8k^2\epsilon^2/\binmechboundvar$.



\paragraph{Proof of \cref{lem:D6} (Lemma D.6 in \citet{jin2020provably})}
Equivalently, we can reparametrize the function class $\mathcal{V}$ by letting $A = \beta^2 (\Lambtil)^{-1}$, so we have 
\begin{equation}
    V(.) = \min \{ \max_{a} \w^\top \phi(.,a) + \sqrt{\phi(.,a)^\top A \phi(.,a)}, H  \}
\label{eq:D6-V-eq}
\end{equation}
for $\nn{\w} \leq L$ and $\nn{A} \leq B^2 (\lambda + \Lambtil_\Lambda)^{-1}$. For any two functions $V_1, V_2 \in \mathcal{V}$, let them take the form in \Cref{eq:D6-V-eq} with parameters $(\w_1, A_1)$ and $(\w_2, A_2)$, respectively. Then, since both $\min \{., H \}$ and $\max_a$ are contraction maps, we have:
\newcommand\numberthis{\addtocounter{equation}{1}\tag{\theequation}}

\begin{align*}
    \mathrm{dist}(V_1, V_2) &\leq \sup_{x,a} \left | \left [ \w_1^\top \phi(x,a) + \sqrt{\phi(x,a)^\top A_2 \phi(x,a)} \right ] - \left [ \w_2^\top \phi(x,a) + \sqrt{\phi(x,a)^\top A_2 \phi(x,a)} \right ] \right | \\
    &\leq \sup_{\phi: \nn{\phi} \leq 1} \left | \left[ \w_1^\top \phi + \sqrt{\phi^\top A_2 \phi} \right] - \left[ \w_2^\top \phi + \sqrt{\phi^\top A_2 \phi} \right] \right |\\
    &\leq \sup_{\phi: \nn{\phi} \leq 1} \left | (\w_1 - \w_2)^\top \phi \right | + \sup_{\phi: \nn{\phi} \leq 1} \sqrt{\left | \phi^\top (A_1 - A_2)\phi \right |}\\
    &= \nn{\w_1 - \w_2} + \sqrt{\nn{A_1 - A_2}} \\
    &\leq \nn{\w_1 - \w_2} + \sqrt{\nn{A_1 - A_2}_F} \numberthis 
    \label{eq:D6-dist-eq}
\end{align*}
where the second last inequality follows from the fact that $| \sqrt{x} - \sqrt{y}| \leq \sqrt{| x-y |}$ holds for any $x, y \geq 0$. For matrices, $\nn{.}$ and $\nn{.}_F$ denote the matrix operator norm and Frobenius norm respectively.

Let $\mathcal{C}_{\w}$ be an $\epsilon/2$-cover of $\{ \w \in \mathbb{R}^d | \nn{\w} \leq L  \}$ with respect to the $2$-norm, and $\mathcal{C}_A$ be an $\epsilon^2/4$-cover of $\{ A \in \mathbb{R}^{d \times d} | \nn{A}_F \leq d^{1/2} B^2 \binmechboundvar^{-1}\}$ with respect to the Frobenius norm. By \Cref{lem:D5}, we know:
\begin{equation*}
    |\mathcal{C}_{\w}| \leq (1 + 4L/\epsilon)^d, \quad |\mathcal{C}_A| \leq [1 + 8d^{1/2}B^2/(\lambtil_\Lambda \epsilon^2)]^{d^2}
\end{equation*}
By \Cref{eq:D6-dist-eq}, for any $V_1 \in \mathcal{V}$, there exists $\w_2 \in \mathcal{C}_{\w}$ and $A_2 \in \mathcal{C}_A$ such that $V_2$ parametrized by $(\w_2, A_2)$ satisfies $\mathrm{dist}(V_1, V_2) \leq \epsilon$. Hence, it holds that $\mathcal{N}_{\epsilon} \leq | \mathcal{C}_{\w}| \dot | \mathcal{C}_A|$, which gives:
\begin{equation*}
    \log \mathcal{N}_{\epsilon} \leq \log | \mathcal{C}_{\w}| + \log |\mathcal{C}_A | \leq d \log(1 + 4L/\epsilon) + d^2 \log [ 1 + 8 d^{1/2} B^2 / ( \lambtil_\Lambda \epsilon^2)]
\end{equation*}
This concludes the proof. 

\paragraph{Proof of \cref{lem:B3} (Lemma B.3 in \citet{jin2020provably})}
For all $(k,h) \in [K] \times [H]$, by \Cref{lem:B2}, we have $\nn{w_h^k} \leq U_K$. In addition, by construction of $\Lambtil_h^{\tilde{k}}$, the minimum eigenvalue of $\Lambtil_h^{\tilde{k}}$ is lowered bounded by $\lambtil_\Lambda$. Thus, by combining \Cref{lem:D4} and \Cref{lem:D6}, with probability at least $1 - \frac{p}{6H}, $we have for any $k > 1$ that:
\begin{align}
    &\nn{ \sum_{i=1}^{\tilde{k} - 1} \phi_h^i [V_{h+1}^{\tilde{k}}(x_{h+1}^i) - \bbP_h V_{h+1}^{\tilde{k}} (x_h^i, a_h^i)] }^2_{(\Lambtil_h^{\tilde{k}})^{-1}}  \\
    \quad&\leq 4H^2 \left[ \frac{d}{2} \log\left( \frac{k + \lambtil_\Lambda}{ \lambtil_\Lambda}  \right) + d \log \left(1 + \frac{4U_K}{\epsilon_0 } \right) + d^2 \log \left( 1 + \frac{8d^{1/2} \beta^2 }{\epsilon_0^2 \lambtil_\Lambda} \right) + \log \left( \frac{6H}{p} \right) \right] + \frac{8 k^2 \epsilon_0^2}{\lambtil_\Lambda}
\label{eq:13-B3}
\end{align}
with $\epsilon_0 = \frac{dH}{k}\sqrt{ \lambtil_\Lambda}$.
Notice that we choose the hyperparameter $\beta = \betabound$. 
We have:
\begin{align*}
    &\quad \nn{ \sum_{i=1}^{\tilde{k} - 1} \phi_h^i [V_{h+1}^{\tilde{k}} (x_{h+1}^i) - \bbP_h V_{h+1}^k (x_h^i, a_h^i)] }^2_{(\Lambtil_h^k)^{-1}} \\
    &\leq \underbrace{4 H^2 d^2 \left( 2 + \log \left( 1 + \frac{8 . 25^2 K^2 \sqrt{d} \log(\chi)}{ \lambtil_\Lambda}\right) \right)}_{G_1} 
    + \underbrace{2H^2 d\log \left(1 + \frac{K}{ \lambtil_\Lambda} \right)}_{G_2} \\
    & + \underbrace{4 H^2 d \log \left( 1 + \frac{4 U_K K}{d H \sqrt{ \lambtil_\Lambda}} \right)}_{G_3} + \underbrace{4H^2 \log \left( \frac{6H}{p} \right)}_{G_4}
\end{align*} 
We can bound each term individually:
\begin{itemize}
    \item $G_1 \leq 4H^2 d^2 [2 + \log(1 + 8 \cdot 25^2 K^2 \sqrt{d} \log(\chi))]$\\
    First, we have $\log(1 + 8 \cdot 24^2 K^2 \log^2(\chi)) \leq 9 \cdot 25^2 K^2 \sqrt{d} \log (\chi)$. \\
    
    Notice that if we set $\chi \geq (9 \cdot 25^2 K^2 \sqrt{d})(18 \cdot 25^2 K^2 \sqrt{d}) = 9 \cdot 18 \cdot 25^4  K^4 d$, then we can upper bound $\chi \geq 9 \cdot 25^2 K^2 \sqrt{d} \log(\chi)$.
    Hence, we can write:
    \begin{equation*}
        G_1 \leq 8 H^2 d^2 \log (\chi ) + 4 H^2 d^2 \log (\chi) = 16 H^2 d^2 \log(\chi)
    \end{equation*}
    \item We have $G_2 \leq 2H^2 d^2 \log(2K) \leq 2H^2 d^2 \log(\chi)$ since $\chi \geq 2K$.
    \item We have $G_3 = 4H^2 d \log \left( 1 + \frac{4 U_K K }{dH \sqrt{\binmechboundvar}} \right) \leq 4 H^2 d^2 \log(5U_K K) \leq 4H^2 d^2 \log(\chi)$ since $\chi \geq 5 U_K K$
    \item We have $G_4 \leq 4H^2 d^2 \log(\chi)$
\end{itemize}
Therefore, 
\begin{equation*}
    \left\|
    \sum_{i=1}^{\tilde{k}-1 }\phi_h^i
    \left[
    V_{h+1}^k (x_{h+1}^i) 
    -
    \bbP_h V_{h+1}^k (x_h^i, a_h^i)
    \right]
    \right\|^2_{(\Lambtil_h^{\tilde{k}})^{-1}}
    \leq 26 d^2 H^2 \log (\chi)
\end{equation*}

\paragraph{Proof of \cref{lem:B4} (Lemma B.4 in \citet{jin2020provably})}
By \Cref{prop:linear-func} and the Bellman equation, we have that for any tuple $(x,a,h) \in \mathcal{S} \times \mathcal{A} \times [H]$:
\begin{equation*}
    Q^{\pi}_h (x,a) := \langle \phi(x,a), \w_h^{\pi} \rangle = (r_h + \bbP_h V^{\pi}_{h+1})(x,a)
\end{equation*}

For any $k\in[K]$, the action-value function in \cref{alg:privrl} is defined as
\begin{align*}
    Q_h^k(x,a) = \vdot{\phi(x,a), \w_h^{\tilde{k}}} 
\end{align*}
where $ \w_h^{\tilde{k}} = \pp{\Lambtil\ii{\tilde{k}}{h}}^{-1}\widetilde{\y}_h^{\tilde{k}}$ and 
%
\begin{align*}
    \Lambtil\ii{{\tilde{k}}}{h} &= 2\binmechboundvar \I + \M_h^{\tilde{k}  } + \sum_{i=1}^{\tilde{k}-1} \phi\pp{x\ii{i}{h}, a\ii{i}{h}}\phi\pp{x\ii{i}{h}, a\ii{i}{h}}^\top  \\
    \ytil_h^{\tilde{k}} &= \sum_{i=1}^{\tilde{k}-1}\phib{x\ii{i }{h}, a\ii{i}{h}} 
    \left( 
    r_h(x\ii{i }{h}, a\ii{i}{h}) + 
    V_{h+1}^{\tilde{k}}(x_{h+1}^i)
                \right) + \etabf_h^{\tilde{k}}
\end{align*}
Note that $\tilde{k} \leq k$ is the last update episode for any $h\in[H]$ before episode $k$ and 
under event $\mathcal{E}$, the minimum eigenvalue of $\Lambtil\ii{k}{h}$ is $\binmechboundvar$. This means that
\begin{align}
\label{eq:norminequality}
  \|v\|_{(\Lambtil\ii{k}{h})^{-1}}\leq
\|v\|_{((\binmechboundvar)\I)^{-1}}
=\sqrt{v^\top (\binmechboundvar\I)^{-1} v}
\leq 
\sqrt{\frac{1}{ \binmechboundvar}}\|v\|  
\end{align}
Hence, we begin by decomposing the term $\w^{\tilde{k}}_h - \w^{\pi}_h$ as
\begin{align*}
    &\quad \w^{\tilde{k}}_h - \w^{\pi}_h\\
    &= (\Lambtil^{\tilde{k}}_h)^{-1} \pp{\sum_{i=1}^{{\tilde{k}}-1} \phi^i_h [r^i_h + V^{\tilde{k}}_{h+1}(x^i_{h+1})] + \etabf_h^{\tilde{k}}} - \w^{\pi}_h\\
     &= (\Lambtil^{\tilde{k}}_h)^{-1} \pp{\sum_{i=1}^{{\tilde{k}}-1} \phi^i_h [r^i_h + V^{\tilde{k}}_{h+1}(x^i_{h+1})] } - \w^{\pi}_h + \pp{\Lambtil_h^{\tilde{k}}}^{-1} \etabf_h^{\tilde{k}} \\
    &= (\Lambtil^{\tilde{k}}_h)^{-1} \left( \sum_{i=1}^{\tilde{k} - 1} \phi_h^i r_h^i + \sum_{i=1}^{\tilde{k} - 1} \phi_h^i V_{h+1}^{\tilde{k}}(x_{h+1}^i) + \phi_h^i \bbP_h V_{h+1}^{\pi} (x_h^i, a_h^i) - \phi_h^i \bbP_h V_{h+1}^{\pi}(x_h^i, a_h^i) \right) \\
    &\quad - \w^{\pi}_h + \pp{\Lambtil_h^{\tilde{k}}}^{-1} \etabf_h^{\tilde{k}}\\
    &= (\Lambtil^{\tilde{k}}_h)^{-1} \left(\sum_{i=1}^{\tilde{k} - 1} \phi_h^i \left(r_h^i + \bbP_h V_{h+1}^{\pi}(x_h^i, a_h^i) \right) + \sum_{i=1}^{\tilde{k} - 1} \phi_h^i \left( V_{h+1}^{\tilde{k}}(x_{h+1}^i) - \bbP_h V_{h+1}^{\pi}(x_h^i, a_h^i) \right) \right)\\
    &\quad - \w_h^{\pi} + \pp{\Lambtil_h^{\tilde{k}}}^{-1} \etabf_h^{\tilde{k}}
\end{align*}
By definition, we can replace $\vdot{\phi_h^i, \w_h^{\pi}} = r_h^i + \bbP_h V_{h+1}^{\pi} (x_h^i, a_h^i)$. Then, we can continue expanding the equality above as:

\begin{align*}
    &\quad \w^{\tilde{k}}_h - \w^{\pi}_h\\
    &= (\Lambtil^{\tilde{k}}_h)^{-1} \left(\sum_{i=1}^{\tilde{k}-1} \phi_h^i \langle \phi_h^i, \w_h^{\pi} \rangle + \sum_{i=1}^{\tilde{k} - 1} \phi_h^i \left( V_{h+1}^{\tilde{k}}(x_{h+1}^i) - \bbP_h V_{h+1}^{\pi}(x_h^i, a_h^i) \right)  \right) - \w_h^{\pi} + \pp{\Lambtil_h^{\tilde{k}}}^{-1} \etabf_h^{\tilde{k}}\\
    &= (\Lambtil^{\tilde{k}}_h)^{-1} \left(\sum_{i=1}^{\tilde{k}-1} \phi_h^i \langle \phi_h^i, \w_h^{\pi} \rangle - \Lambtil_h^{\tilde{k}} \w_h^{\pi} + \sum_{i=1}^{\tilde{k} - 1} \phi_h^i \left( V_{h+1}^{\tilde{k}}(x_{h+1}^i) - \bbP_h V_{h+1}^{\pi}(x_h^i, a_h^i) \right)  \right) + \pp{\Lambtil_h^{\tilde{k}}}^{-1} \etabf_h^{\tilde{k}}\\
\end{align*}
Then, plugging in for the definition of $ \Lambtil_h^{\tilde{k}}$ we get 
\begin{align*}
    &= (\Lambtil^{\tilde{k}}_h)^{-1} \left(\sum_{i=1}^{\tilde{k}-1} \phi_h^i \langle \phi_h^i, \w_h^{\pi} \rangle - \left(2\binmechboundvar \I + \M_h^{\tilde{k}} +  \sum_{i=1}^{\tilde{k}-1} \phi_h^i {\phi_h^i}^\top  \right) \w_h^{\pi} \right ) \\
    &\quad + (\Lambtil^{\tilde{k}}_h)^{-1} \left ( \sum_{i=1}^{\tilde{k} - 1} \phi_h^i \left( V_{h+1}^{\tilde{k}}(x_{h+1}^i) - \bbP_h V_{h+1}^{\pi}(x_h^i, a_h^i) \right)  \right) 
    + \pp{\Lambtil_h^{\tilde{k}}}^{-1} \etabf_h^{\tilde{k}}\\
    &= (\Lambtil^{\tilde{k}}_h)^{-1} \left\{ \left(-2\binmechboundvar \I - \M_h^{\tilde{k}} \right) \w_h^{\pi}  + \sum_{i=1}^{{\tilde{k}}-1} \phi^i_h [V^{\tilde{k}}_{h+1}(x^i_{h+1}) - \bbP_h V^{\pi}_{h+1}(x^i_h, a^i_h)] \right\} + \pp{\Lambtil_h^{\tilde{k}}}^{-1} \etabf_h^{\tilde{k}} \\
    &= \underbrace{-(\Lambtil_{h}^{\tilde{k}})^{-1} \left( 2\lambtil_{\Lambda}\I + \M_h^{\tilde{k}} \right)\w_h^{\pi}}_{\textbf{q}_1} + \underbrace{(\Lambtil^{\tilde{k}}_h)^{-1} \sum_{i=1}^{{\tilde{k}}-1} \phi^i_h [V^{\tilde{k}}_{h+1} (x^i_{h+1}) - \bbP_h V^{\tilde{k}}_{h+1}(x^i_h, a^i_h)]}_{\textbf{q}_2} + \\
    &\quad + \underbrace{(\Lambtil_h^{\tilde{k}})^{-1} \sum_{i=1}^{{\tilde{k}}-1} \phi^i_h \bbP_h (V^{\tilde{k}}_{h+1} - V^{\pi}_{h+1}) (x^i_h, a^i_h)}_{\textbf{q}_3}  + \underbrace{\pp{\Lambtil_h^{\tilde{k}}}^{-1} \etabf_h^{\tilde{k}}}_{\textbf{q}_4}
\end{align*}

We proceed to bound each term in the right hand side individually. For the first term, we have:
\begin{align*}
    |\langle \phi(x,a), \textbf{q}_1 \rangle| 
    &= 
    \vdot{
    \phi(x,a), 
    (\Lambtil_h^{\tilde{k}})^{-1}\left( 2\lambtil_{\Lambda}\I + \M_h^{\tilde{k}} \right)\w_h^{\pi}
    } \\
    &= 
    \vdot{
    \phi(x,a)(\Lambtil_h^{\tilde{k}})^{-1/2}, 
    (\Lambtil_h^{\tilde{k}})^{-1/2}\left( 2\lambtil_{\Lambda}\I + \M_h^{\tilde{k}} \right)\w_h^{\pi}
    } \\
    &\leq 
    \|\phi(x,a)\|_{(\Lambtil_h^{\tilde{k}})^{-1}}
    \left\|
    \left(
    2\lambtil_{\Lambda}\I + \M_h^{\tilde{k}}
    \right)
    \w_h^{\pi}\right\|_{(\Lambtil_h^{\tilde{k}})^{-1}}\\
    &\leq 
    \|\phi(x,a)\|_{(\Lambtil_h^{\tilde{k}})^{-1}}
    \left\|
    \left(
    \lambtil_{\Lambda}\I 
    \right)
    \w_h^{\pi}\right\|_{(\Lambtil_h^{\tilde{k}})^{-1}}\\
    &\leq \frac{\binmechboundvar}{\sqrt{ \binmechboundvar}}   \nn{w_h^{\pi}} \|\phi(x,a)\|_{(\Lambtil_h^{\tilde{k}})^{-1}} \tag{by \cref{eq:norminequality}} \\
    &= 2 H \sqrt{d\binmechboundvar} \nn{\phi(x,a)}_{(\Lambtil_h^{\tilde{k}})^{-1}} \tag{by \Cref{lem:B1}}\\
    %
\end{align*}

As a straightforward application of \Cref{lem:B3}, we can bound the second term as such:
\begin{align*}
    |\langle \phi(x,a), \textbf{q}_2 \rangle|&\leq 
   6 d H \sqrt{\log(\chi)} 
    \sqrt{\phi(x,a)^\top (\Lambtil^{\tilde{k}}_h)^{-1} \phi(x,a)} \\
    &= 6dH \sqrt{\log(\chi)} \nn{\phi(x,a)}_{(\Lambtil_h^{\tilde{k}})^{-1}} 
\end{align*}
\color{black}

For the third term, by definition of linear MDP, we have:

\begin{align*}
    \langle \phi(x,a), \textbf{q}_3 \rangle &= \left \langle \phi(x,a), (\Lambtil_h^{\tilde{k}})^{-1} \sum_{i=1}^{\tilde{k}-1} \phi^i_h \bbP_h(V^{\tilde{k}}_{h+1} - V^{\pi}_{h+1})(x^i_h, a^i_h) \right \rangle \\
    &= \left \langle \phi(x,a), (\Lambtil^{\tilde{k}}_h)^{-1} \sum_{i=1}^{\tilde{k}-1} \phi^i_h {\phi^i_t}^\top \int (V^{\tilde{k}}_{h+1} - V^{\pi}_{h+1})(x') d\mu_h(x') \right \rangle\\
    &= \underbrace{\left \langle \phi(x,a), \int (V^{\tilde{k}}_{h+1} - V^{\pi}_{h+1})(x')d\mu_h(x') \right \rangle}_{p_1}\\
    &\quad \underbrace{-  \left \langle \phi(x,a), (\Lambtil^{\tilde{k}}_h)^{-1} \left( 2\binmechboundvar \I + \M_h^{\tilde{k}} \right) \int (V^{\tilde{k}}_{h+1} - V^{\pi}_{h+1})(x')d\mu_h(x') \right \rangle}_{p_2}
\end{align*}
where, by assumption, we have
\begin{align*}
    p_1 &= \bbP_h(V^{\tilde{k}}_{h+1} - V^{\pi}_{h+1})(x,a)
\end{align*}
and similar to $\textbf{q}_1$, we can bound $|p_2|$ as follows:
\begin{align*}
    |p_2| &\leq \sqrt{\phi(x,a)^\top (\Lambtil^{\tilde{k}}_h)^{-1} \phi(x,a)} \nn{(\Lambtil_h^{\tilde{k}})^{-1/2}} \nn{(2\binmechboundvar) \I + \M_h^{\tilde{k}}} 2H \sqrt{d}\\
    &\leq \frac{1}{\sqrt{\binmechboundvar}} (\lambtil_{\Lambda}) 2H \sqrt{d} \sqrt{\phi(x,a)^\top (\Lambtil^{\tilde{k}}_h)^{-1} \phi(x,a)} \\
    &\leq  2H \sqrt{d\binmechboundvar} \nn{\phi(x,a)}_{(\Lambtil_h^{\tilde{k}})^{-1}} 
\end{align*}
\color{black}

For the fourth term, we have:
\begin{align*}
    |\langle \phi(x,a), \textbf{q}_4\rangle|
    &\leq \left | \left\langle \phi(x,a), \pp{\Lambtil_h^{\tilde{k}}}^{-1} \etabf_h^{\tilde{k}} \right\rangle \right | \\
    &= \left | \phi(x,a)^\top (\Lambtil_h^{\tilde{k}})^{-1/2} (\Lambtil_h^{\tilde{k}})^{-1/2} \etabf_h^{\tilde{k}} \right | \\
    &= \left | \left((\Lambtil_h^{\tilde{k}})^{-1/2} \phi(x,a) \right)^\top \left((\Lambtil_h^{\tilde{k}})^{-1/2} \etabf_h^{\tilde{k}} \right) \right | \\
    &\leq \nn{(\Lambtil_h^{\tilde{k}})^{-1/2} \phi(x,a)} \nn{(\Lambtil_h^{\tilde{k}})^{-1/2} \etabf_h^{\tilde{k}}}\\
    &= \nn{(\Lambtil_h^{\tilde{k}})^{-1/2} \etabf_h^{\tilde{k}}} \sqrt{\phi(x,a)^\top (\Lambtil_h^{\tilde{k}})^{-1}\phi(x,a)} \\
    &= \nn{\etabf_h^{\tilde{k}}}_{(\Lambtil_h^{\tilde{k}})^{-1}}\sqrt{\phi(x,a)^\top (\Lambtil_h^{\tilde{k}})^{-1}\phi(x,a)} \\
    &\leq \frac{\nn{\etabf_h^{\tilde{k}}}}{\sqrt{\binmechboundvar}} \sqrt{\phi(x,a)^\top (\Lambtil_h^{\tilde{k}})^{-1}\phi(x,a)} \\
    &\leq \frac{\gaumechboundvar}{ \sqrt{ \binmechboundvar}}  \nn{\phi(x,a)}_{(\Lambtil_h^{\tilde{k}})^{-1}} 
\end{align*}
Finally, since we have $\langle \phi(x,a), \w_h^{\tilde{k}} \rangle - Q_h^{\pi}(x,a) = \langle \phi(x,a), \w_h^{\tilde{k}} - \w_h^{\pi} \rangle = \langle \phi(x,a), \textbf{q}_1 + \textbf{q}_2 + \textbf{q}_3 + \textbf{q}_4 \rangle$, we can write:

    

\begin{align*}
    &\left | \langle \phi(x,a), \w_h^{\tilde{k}} \rangle - Q_h^{\pi}(x,a) - \bbP_h(V^{\tilde{k}}_{h+1} - V^{\pi}_{h+1})(x,a)\right| \\
    &\leq \left( 2  H \sqrt{d\binmechboundvar}  + 6 d H \sqrt{\log(\chi)} + 2 H \sqrt{d\binmechboundvar} + \frac{\gaumechboundvar}{ \sqrt{ \binmechboundvar}}  \right) \nn{\phi(x,a)}_{(\Lambtil_h^{\tilde{k}})^{-1}}
\end{align*}

Hence, for the inequality to hold, we need to set \begin{equation}
\label{eq:beta-condition}
    \beta \geq  2  H \sqrt{d\binmechboundvar}  + 6 d H \sqrt{\log(\chi)} + 2 H \sqrt{d\binmechboundvar} + \frac{\gaumechboundvar}{ \sqrt{ \binmechboundvar}} 
\end{equation}


Observe that the RHS of inequality \eqref{eq:beta-condition} becomes:
\begin{align*}
    \mathrm{\text{RHS of \cref{eq:beta-condition}}} &=  2  H \sqrt{d\binmechboundvar}  + 6 d H \sqrt{\log(\chi)} + 2 H \sqrt{d\binmechboundvar} + \frac{\gaumechboundvar}{ \sqrt{ \binmechboundvar}} \\
    &\leq 4  H \sqrt{d\binmechboundvar}  + 6 d H \sqrt{\log(\chi)} + \frac{\gaumechboundvar}{ \sqrt{ \binmechboundvar}} 
\end{align*}

Finally, we upper bound the ratio $\pp{\gaumechboundvar}/\sqrt{\binmechboundvar}$. Recall that $\gaumechboundvar=\gaumechbound$ and $\binmechboundvar=\binmechbound$, therefore 
\begin{align*}
   \frac{\gaumechboundvar}{\sqrt{\binmechboundvar}}
   = \frac{\sqrt{2}\pp{\frac{dH^2}{\log 2} \log\pp{1 + \frac{K}{\binmechboundvar d}}}\sqrt{\log\pp{3KH/p}}}
   {\sqrt{\log\pp{K} \pp{6\sqrt{d+1} + \log\pp{3KH/p}}}}
\end{align*}

Observe that we can upper bound the last term by $H^2\sqrt{d \binmechboundvar \log(\chi)}$. 
Therefore, it suffices to set $\beta = 5 H^2 \sqrt{d\binmechboundvar \log(\chi)} + 6dH \sqrt{\log(\chi)}$.

\paragraph{Proof of \cref{lem:B5} (Lemma B.5 in \citet{jin2020provably})}
We will prove this lemma by induction.
Base case: At the last step $H$, the statement is true because $Q_H^k (x,a ) \geq Q^*_H (x,a)$. Since the value function at step $H+1$ is zero, by \Cref{lem:B4}, we have:
\begin{equation*}
    | \langle \phi(x,a), \w_H^{\tilde{k}} \rangle - Q_H^*(x,a) | \leq \beta \sqrt{\phi(x,a)^\top (\Lambtil_H^{\tilde{k}})^{-1} \phi(x,a)}
\end{equation*}
Hence, we have:
\begin{equation*}
    Q_H^* (x,a) \leq \min \{ \langle \phi(x,a), \w_H^{\tilde{k}} \rangle + \beta \sqrt{\phi(x,a)^\top (\Lambtil_H^{\tilde{k}})^{-1} \phi(x,a)} , H\} = Q_H^{\tilde{k}}(x,a).
\end{equation*}
Induction hypothesis: Suppose the statement is true at step $h+1$. Consider step $h$. By \Cref{lem:B4}, we have:
\begin{equation*}
    | \langle  \phi(x,a), \w_h^{\tilde{k}} \rangle |  - Q_h^* (x,a) - \bbP_h (V_{h+1}^{\tilde{k}} - V^*_{h+1})(x,a) \leq \beta \sqrt{\phi(x,a)^\top (\Lambtil_h^{\tilde{k}})^{-1} \phi(x,a)}
\end{equation*}
By the induction assumption that $\bbP_h (V_{h+1}^{\tilde{k}} - V^*_{h+1}) (x,a) \geq 0$. we have:
\begin{equation*}
    Q_h^* (x.a) \leq \min \{ \langle \phi(x,a), \w_h^{\tilde{k}}  \rangle + \beta \sqrt{\phi(x,a)^\top (\Lambtil_h^{\tilde{k}})^{-1} \phi(x,a)} , H  \} = Q_h^k(x,a)
\end{equation*}
This concludes the proof. 

\section{Switching Cost Analysis}
\label{sec:switching-cost-appendix}



First, we give an upper bound on the determinant of $\Lambtil_h^k$. 
\begin{lemma}[Similar to Lemma C.1 in \cite{wang2021provably}] Let $\{ \Lambtil_h^k, (k,h) \in [K] \times [H] \}$ be as defined in \Cref{alg:privrl}. Then for all $h \in [H]$ and $k \in [K]$, we have $\det \left( \Lambtil_h^k \right) \leq (\binmechboundvar + (k-1)/d)^d$
\label{lem:C1}
\end{lemma}

\begin{proof}
We have 
\begin{align*}
    \Tr(\Lambtil_h^k) &= \Tr(2\lambtil \I + \M_h^k) +  \sum_{i=1}^{k-1} \Tr(\phi(x_h^i, a_h^i) \phi(x_h^i, a_h^i)\top)\\
    &= \binmechboundvar d + \sum_{i=1}^{k-1} \nn{\phi(x_h^i, a_h^i)}_2^2 \leq \binmechboundvar d + k - 1
\end{align*}
where the last inequality is because we assume $\nn{\phi(x,a)} \leq 1$. Since $\Lambtil_h^k$ is PSD, by AM-GM, we have
\begin{align*}
    \det(\Lambtil_h^k) \leq \left( \frac{\Tr(\Lambtil_h^k)}{d} \right)^d \leq \left(\binmechboundvar + \frac{k-1}{d} \right)^d
\end{align*}
\end{proof}

Next, we provides a determinant-based upper bound for the ratio between the norms $\nn{\cdot}_A$ and $\nn{\cdot}_B$, where $A \succeq B$.
\begin{lemma} [Lemma 12 in \cite{abbasi2011improved}]
Suppose $A, B \in \mathbb{R}^{d \times d}$ are two PSD matrices such that $A \succeq B$, then for any $x \in \mathbb{R}^d$, we have $\nn{x}_A \leq \nn{x}_B \cdot \sqrt{\det(A)/\det(B)}$
\end{lemma}

Finally, we can derive the switching cost of \Cref{alg:privrl} in the following lemma:

\begin{lemma}[Similar to Lemma C.3 in \cite{wang2021provably}] Condition on the event that $\nn{\M_h^k} \leq \binmechboundvar$ for all $h, k \in [H] \times [K]$. 
For $C=2$ and $\binmechboundvar > 0$, the global switching cost of \Cref{alg:privrl} is bounded by: 
\begin{align*}
    \switchcountvar \leq \frac{dH}{\log 2 } \log \left(1 + \frac{K}{\binmechboundvar d}  \right) 
\end{align*}
\end{lemma}

\begin{proof}
Let $\{k_1, k_2, \cdots, k_{\switchcountvar} \}$ be the episodes where \Cref{alg:privrl} updates the policy, and let $k_0 = 0$. Then, by the update condition on line \eqref{lin:update}, for each $i \in [\switchcountvar]$, there exists at least one $h \in [H]$ such that
\begin{align*}
    \det (\Lambtil_h^{k_i}) > 2 \det (\Lambtil_h^{k_{i-1}})
\end{align*}
By the definition of $\Lambtil_h^k$, we know that $\Lambtil_h^{i_1} \succeq \Lambtil_h^{i_2}$ for all $i_1 \geq i_2$ and $h \in [H]$. Hence, we have
\begin{align*}
    \prod_{h=1}^H \det(\Lambtil_h^{k_i}) > 2 \prod_{h=1}^H \det(\Lambtil_h^{k_{i-1}})
\end{align*}
We can recursively apply the inequality above to all $i \in [\switchcountvar]$ and get
\begin{align*}
    \prod_{h=1}^H \det(\Lambtil_h^{k_{\switchcountvar}}) > 2^{\switchcountvar} \cdot \prod_{h=1}^H \det(\Lambtil_h^1) = 2^{\switchcountvar} \binmechboundvar^{dH}
\end{align*}
as we initialize $\Lambtil_h^1 = 2\binmechboundvar \I$. 

Also, by \Cref{lem:C1}, we have
\begin{align*}
    \prod_{h=1}^H \det \left( \Lambtil_h^{k_{\switchcountvar}} \right) \leq \prod_{h=1}^H \det (\Lambtil_h^K) \leq \left(\binmechboundvar + \frac{K}{d} \right)^{dH}
\end{align*}
Therefore, we can combine the two inequalities above and get
\begin{align*}
    \switchcountvar \leq \frac{dH}{\log 2} \log \left( 1 + \frac{K}{\binmechboundvar d} \right)
\end{align*}
\end{proof}

\section{Auxiliary Results}
\begin{claim}[Concentration inequalities in \citep{tao2012topics}] Let $M \in \mathbb{R}^{d \times d}$ be a symmetric matrix where each of its entries $M_{i,j} = M_{j,u} \sim \cN(0,1)$ for any $1 \leq i \leq j \leq d$. Then, for any $\alpha > 0$, $\bbP(\nn{M}_{op} \geq 4\sqrt{d} + 2 \log \left( \frac{1}{\alpha} \right) ) \leq \alpha$, where $\nn{M}_{op}$ is the operator norm of a matrix associated to the norm $\nn{\cdot}_2$.
\end{claim}

\begin{claim}[Corollary to Lemma 1 in \citep{Laurent2005AdaptiveEO}] If $U \sim \chi^2(d)$ and $\alpha \in (0,1)$:
\begin{align*}
    \bbP \left( U \geq d + 2 \sqrt{d \log \left( \frac{1}{\alpha} \right) + 2 \log \left( \frac{1}{\alpha} \right)} \right) \leq \alpha\\
    \bbP \left( U \leq d - 2\sqrt{d \log \left( \frac{1}{\alpha} \right)} \right) \leq \alpha
\end{align*}
As a consequence of the first inequality, we also have that for any vector $v \in \mathbb{R}^d$ drawn from a $d-$dimensional Gaussian distribution $\cN(0, \I_{d \times d})$, then $\bbP \left( \nn{v}_2 > \sqrt{d} + 2 \sqrt{\log \left( \frac{1}{\alpha} \right)}  \right) \leq \alpha$.
\end{claim}

\begin{proof}
By definition of Laplace distribution, for each $i \in [d]$, we have
\begin{equation*}
    \bbP \left( |v_i| > \log \left(\frac{d}{\alpha} \right) \right) \leq \frac{\alpha}{d}
\end{equation*}
Hence, in the union event, with probability at least $1 - \alpha$, we have $\nn{v}_2 \leq \sqrt{d} \log \left( \frac{d}{\alpha} \right)$.
\end{proof}

\begin{claim}[Theorem 7.8 in \citep{zhang11matrix}] For two positive definite matrices (PSD) $A, B \in \mathbb{R}^{d \times d}$, we write $A \succeq B$ to denote that $A - B$ is PSD. Then, if $A \succeq B \succeq 0$, we have:
\begin{itemize}
    \item $\mathrm{rank}(A) \geq \mathrm{rank}(B)$
    \item $\det(A) \geq \det(B)$
    \item $B^{-1} \succeq A^{-1}$ if $A$ and $B$ are non-singular.
\end{itemize}
\end{claim}

\begin{claim}[Lemma 12 in \citep{abbasi2011improved}] Supposed $A, B \in \mathbb{R}^{d \times d}$ are two PSD matrices such that $A \succeq B$. Then, for any $x \in \mathbb{R}^{d}$, we have $\nn{x}_A \leq \nn{x}_B \sqrt{\frac{\det(A)}{\det(B)}}$. 

\end{claim}


\end{document}